# Measuring group-separability in geometrical space for evaluation of pattern recognition and embedding algorithms

A. Acevedo, S. Ciucci, MJ. Kuo, C. Durán and CV. Cannistraci

**Abstract** — Evaluating data separation in a geometrical space is fundamental for pattern recognition. A plethora of dimensionality reduction (DR) algorithms have been developed in order to reveal the emergence of geometrical patterns in a low dimensional visible representation space, in which high-dimensional samples similarities are approximated by geometrical distances. However, statistical measures to evaluate directly in the low dimensional geometrical space the sample group separability attaiend by these DR algorithms are missing. Certainly, these separability measures could be used both to compare algorithms performance and to tune algorithms parameters. Here, we propose three statistical measures (named as PSI-ROC, PSI-PR, and PSI-P) that have origin from the Projection Separability (PS) rationale introduced in this study, which is expressly designed to assess group separability of data samples in a geometrical space. Traditional cluster validity indices (CVIs) might be applied in this context but they show limitations because they are not specifically tailored for DR. Our PS measures are compared to six baseline cluster validity indices, using five non-linear datasets and six different DR algorithms. The results provide clear evidence that statistical-based measures based on PS rationale are more accurate than CVIs and can be adopted to control the tuning of parameter-dependent DR algorithms.

## 1 Introduction

One of the main current problems in data science, machine learning and pattern recognition is to develop appropriate visual representations of complex data [1]–[4]. In the case of high-dimensional datasets, the process of discovering patterns is further complicated by the fact that the data often cannot be immediately visualized to determine the similarity and separability of groups of samples. Thus, the development of different techniques for dimensionality reduction (DR) or data embedding have attracted much attention [5]. These techniques produce a low dimensional representation where the geometrical distances between samples preserve the similarities of the high dimensional data together with their relevant structure [6].

Multiple DR algorithms have been developed such as: PCA [7], [8], MDS [9], t-SNE [10], [11], Isomap [12], and MCE [2]. Despite their attempt to preserve the original data structure, many of these algorithms may partially fail, and a measure or index that can quantify and evaluate the geometrical separation of the groups of samples, according to the geometrical patterns revealed by these techniques is needed. We can think of clustering [13] as a similar scenario, in which clustering validity indices (CVIs) help to identify the correct sample grouping based on clustering results [14], [15]. Hence, CVIs might be naïvly applied in order to validate the results of DR techniques. Therefore, we considered in this study six CVIs most commonly used in literature such as: Dunn index (DN) [16] that relies on the distances among clusters and their diameters; Davies-Bouidlin index (DB) [17] based on the idea that for a good partition inter-cluster separation as well as intra-cluster homogeneity and compactness should be high; Calinski-Harabasz index (CH) [18] based on the average between-cluster means and within-cluster sum of squares; Silhouette index (SH) [19] that validates the clustering performance based on the pairwise difference of between-cluster and within-cluster distances; Bezdek index (BZ) [20] a variation of the Dunn index created by Bezdek and Pal, and described in [21]; Thornton's separability (TH) index [22], which calculates the average number of instances that share the same class label as their nearest neighbours.

Our results show that the use of the CVI techniques is not suitable for the validation of DR methods, because these indices: i) are not tailored for comparison of results across different dimensions (for instance the result of an index applied in 2 dimensions is not directly comparable with the result of the same index applied in 3 dimensions); ii) do not always address properly the problem of group separability in a geometrical space.

Group separability aims to evaluate in a geometrical space the extent to which groups (classes or populations) that are partially overlapping [21],[22], and display a pattern of separation between them. However, in high-dimensional settings, several groups may exist in different subspaces [25], [26], one of the consecuences of the so-called

- *A. Acevedo, C. Duran, S. Ciucci, MJ. Kuo are with the Biomedical Cybernetics Group, Biotechnology Center (BIOTEC), Center for Molecular and Cellular Bioengineering (CMCB), Center for Systems Biology Dresden (CSBD), Technische Universität Dresden, Dresden, Germany. E-mail: aldo.acevedo.toledo@gmal.com, claudio134@gmail.com, saraciuccipg@gmail.com, juo.ming.ju@gmail.com.*
- *CV. Cannistraci is with the Biomedical Cybernetics Group, Biotechnology Center (BIOTEC), Center for Molecular and Cellular Bioengineering (CMCB), Center for Systems Biology Dresden (CSBD), Technische Universität Dresden, Dresden, Germany and with the Complex Network Intelligence center, Tsinghua Laboratory of Brain and Intelligence, Tsinghua University, Beijing, China. E-mail: kalokagathos.agon@gmail.com.*

"curse of dimensionality" [27]. Thus, a suitable measure of group separability is required in order to judge the group separability provided by different dimensionality reduction methods. Here we propose a novel rationale named Projection Separability (PS), which is specifically designed to assess group separability of data samples in a geometrical space such as in the case of dimensionality reduction analyses based on embedding algorithms. Based on this new rationale, we implemented three statistical separability measures (which we called as Projection Separability Indices, PSIs) based on geometrically projecting the samples (points) of two different groups on the line that connects their centroids, and then repeating this procedure for all pairs of groups for which separability evaluation is desired.

We have compared the state of the art CVIs with our PSIs, and showed the effectiveness of PSIs to validate the group separability given by six DR techniques on five high-dimensional datasets (one syntetic and four real data). Finally, we propose the exploitation of PSIs for: I) automatic validation and identification of the best dimensionality reduction methods for a certain problem or dataset; II) automatic detection of the best internal tuning parameters in DR methods; and III) automatic detection of the best normalizations associated to a certain dataset in relation to a given DR method.

## 2 Methods

Several state of the art indices, which are all described in this section, were compared to our novel PS approaches. To compare the results of the different indices, five datasets were analyzed by means of diverse dimensionality reduction (DR) methods, considering the first two dimensions of mapping (2D), and in some cases the first three dimension of mapping (3D). In general, DR techniques can efficiently bring down the features/variables space to a much smaller number of dimensions without a significant loss of information [28]. We applied these DR techniques owing to the idea that the indices might pinpoint different best performing DR methods due the diversity of the datasets. The results may change by considering datasets with a small number of samples, with unbalanced sample groups, or even in presence of noisy data. Under these contexts, some indices may report a wrong evaluation of the DR methods, selecting those with an insufficient or inaccurate grouping of the samples.

The applied DR methods are both linear and nonlinear, since linear approaches may miss represent hidden nonlinear relations among the samples in the feature space. The used techinques are the following: Principal Component Analysis (PCA) [7], [8], non-metric Multidimensional Scaling with Sammon Mapping (NMDS) [29], [30], Multidimensional Scaling based on Bray-Curtis dissimilarity (MDSbc) [31], Minimum Curvilinearity Embedding (MCE) [2], Isomap [12], and t-Distributed Stochastic Neighbor Embedding (t-SNE) [10], [11]. Due to the high-dimensionality of the datasets, each method can produce a different partitioning or grouping of the samples. Moreover, linear approaches might suffer absence of sample group separation in the reduced space of representation, which not necessary imply the absence of separation in general, indeed nonlinear techniques for DR may instead prove it. For this reason, the role of the separability indices is to discern and quantitatively assess what DR methods provide the best group separability of the samples. Since this study focuses on the separabilty indices, we do not provide any further explanation of these DR methods (see the corresponding references for more details).

### 2.1 Dataset description

We consider one artificial and four real datasets for the comparison of the proposed PSIs versus multiple traditional CVIs techniques. It is important to mention that all the datasets employed in this study present a non-linear structure. This selection was made with the aim to assess if our PSIs can correctly evaluate how DR methods well detect nonlinearity in data. A completely description of the numbers of samples and features/variables, and the number of classes for each dataset is presented in Table 1.

#### *2.2.1 Tripartite-Swiss-Roll*

The synthetic Tripartite-Swiss-Roll dataset was obtained by discretizing the Swiss-Roll manifold in a three-dimensional space (3D) [12]. It contains 723 points divided in three main groups (clusters) characterized by nonlinear structures. The difficulty stays in the fact that the typical nonlinearity of the Swiss-Roll shape is given by the tripartition of the manifold, further impaired by discontinuity.

#### *2.2.2 Gastric mucosa microbiome*

A first real 16S rRNA gene sequence dataset was generated by Sterbini and colleagues [32]. It is public and available at the NCBI Sequence Read Archive (SRA) (http://www.ncbi.nlm.nih.gov/sra, accession number SRP060417), where all details pertaining the sequencing experimental design are also reported. It contains 24 biopsy specimens of the gastric antrum from 24 individuals who were referred to the Department of Gastroenterology of Gemelli Hospital (Rome) with dyspepsia symptoms (i.e. heartburn, nausea, epigastric pain and discomfort, bloating, and regurgitation). Proton Pump Inhibitors (PPIs) are in fact increasingly being used for the treatment of the most frequent causes of dyspepsia, besides being part of anti-*Helicobacter pylori* treatment regimens [32]. In this dataset, twelve of these individuals had been taking PPIs for at least 12 months, while the others were not being treated (naïve) or had stopped treatment at least 12 months before sample collection. In addition, 9 (4 treated and 5 untreated) were positive for *H. pylori* infection, where *H. pylori* positivity or negativity was determined by histology and rapid urease tests. Metagenomes were obtained by pyrosequencing fragments of the 16S r-RNA gene on the GS Junior platform (454 Life Sciences, Roche Diagnostics). Then the sequence data were processed, replicating the bioinformatics workflow followed by Sterbini et al. [32], in order to obtain the matrix of the bacterial absolute abundance. Therefore, the resulting dataset is composed by a total of 24 samples, 187 features and 3 groups: untreated dyspeptic patients without *H. pylori* infection (HPneg); untreated dyspeptic patients with H. pylori infection (HPpos); and patients treated with Proton Pump Inhibitors (PPI).

#### 2.2.3 Radar signal

A second real dataset was recovered from the *UCI Machine learning repository* (available at http://archive.ics.uci.edu/ml/datasets/Ionosphere). This dataset is widely described and analyzed in [33]. It contains 351 radar signals targeting free electrons in the ionosphere. Shieh and colleagues [34] analyzed this dataset with two labeled groups (good radar signal, and bad radar signal), however, they highlighted that good radar signals are highly similar, and bad radar signals are highly dissimilar. Latterly, Cannistraci *et al.* [35] proposed that the group of bad radar signals might be interpreted as two diverse sub-categories (two different clusters) that are difficult to identify due their high non-linearity (elongated and/or irregular high-dimensional structure). Based on these results, we used three labeled classes: good radar signal, bad radar signal 1, and bad radar signal 2.

In the first stage of the analysis, we detected two samples (radar signals) with exactly the same values across all the features/variables, so we removed one of them in order to avoid problems with the calculation of the dissimilarity matrix employed by NMDS. Therefore, the resulting dataset is composed by a total of 350 samples, 34 features, and 3 groups.

#### 2.2.4 Image Proteomics

We used a proteomic dataset obtained from [2]. It was generated by combining: 1) a dataset from 2D Electrophoresis (2DE) gel images derived from proteomic Cerebrospinal fluid (CSF) samples of peripheral neuropathic patients [36]; and 2) another dataset derived from a neurological study of amyotrophic lateral sclerosis (ALS) patients not affected by neuropathic pain [37]. The resulting dataset contains four main classes: healthy control patients (C) with 8 samples, patients not affected by neuropathic pain (M) with 19 samples, patients without pain (NP) with 8 samples, and patients with a pathological variant with pain (P) with 7 samples. In [2] the authors discovered that four of the patients wihout pain (NP group) developed pain after a clinical follow-up at 6-12 months (or > 1 year). Hence, we considered these four patients in the group pathological variant with pain (P), resulting in 11 samples. The remaining 4 patients of the NP group were instead included in the healthy control patient class (C group), ending up with 12 samples. Therefore, the dataset is composed by a total of 42 samples, 1947 features, and 3 groups.

#### 2.2.5 MNIST

Lastly, we used a well known dataset in the machine learning field called MNIST [38]. This is a large dataset that consists of 28x28 pixel images of handwritten digits. Every image can be thought as a 784-dimensional array, where each value represent each pixel's intensity in gray scale. The different sample groups are numbers between 0 and 9.

Since this is a very large dataset, we randomly selected 30 samples for each digit, resulting in a sub-dataset with 300 samples. Therefore, the dataset is composed by a total of 300 samples, 784 features, and 10 groups.

### 2.3 Group Separability Indices

In this section, we describe all the indices employed in this study and, in order to falicitate their understanding, we have written and explained the mathematical formulation of each of them.

You will notice a major difference between the classical CVIs and our PSIs in the sense that the CVIs do not have an upperbound which indicate the best performance, therefore the same value in two different dimensional space of different size does not necessarily indicate a similar level of separation. Whereas all the PSIs are defined inside a bounded range that is borrowed by the statistics to which they are associated. Thus, their values indicate the same separability regardless of the size of the dimensional space in which it is computed.

The Matlab code to compute all the indices described in this section is available at the following GitHub link: https://github.com/biomedical-cybernetics/projection-separability-indices.

#### 2.3.1 Silhouette index

Silhouette index (SH) [19] is a measure used for fuzzy clustering based on the concept of *silhouette width* defined as follows:

$$SH = \frac{1}{N}\sum_{k=1}^{N} SH(C_k) \quad (1)$$

where $N$ is the number of clusters, and $SH(C_k)$ represents the *silhouette width* for the cluster $C_k$ calculated as:

$$SH(C_k) = \frac{1}{n_k}\sum_{x\in C_k} SH(x) \quad (2)$$

where $n_k$ is the number of samples in $C_k$, and $SH(x)$ is the silhouette width of the sample $x$, and can be written as:

$$SH(x) = \frac{b(x)-a(x)}{\max(a(x),b(x))} \quad (3)$$

Here, $a(x)$ represents the within-cluster mean distance defined by the average distance between $x$ and the rest of the samples belonging to the same cluster. On the other hand, $b(x)$ is the smallest of the mean distances of $x$ to the samples belonging to the others cluster.

The higher the value of SH, the better grouping of the samples is obtained.

#### 2.3.2 Calinski-Harabasz index

Calinski-Harabasz index (CH) [18] is based on the ratio between the overall between-cluster distance and the overall within-cluster distance and it is defined as:

$$\mathrm{CH} = \frac{SS_B}{SS_W} * \frac{T-N}{N-1} \quad (4)$$

where $N$ is the number of clusters, and $T$ is the total number of samples. $SS_B$ is the overall between-cluster distance involving the elements of different clusters, and it is denoted by:

$$SS_B = \sum_{i=1}^{N} T_i \times \|u_i - u\|^2 \qquad (5)$$

$T_i$ is the number of samples in the cluster $i$, $u_i$ is the mean (centroid) of the $i^{th}$cluster, $u$ is an overall mean (overall centroid) of the sample data, and $\|\cdot\|$ denotes the Euclidean distance. In (4), $SS_w$ instead is the overall within-cluster variance, which is calculated as follows:

$$SS_W = \sum_{i=1}^{N} \sum_{x_j \in C_i} \left\|x_j - u_i\right\|^2 \qquad (6)$$

$x_j$ is the data point belonging to the $i^{th}$ cluster represented by $C_i$, $u_i$ represents the centroid of the cluster $i$, and $\|\cdot\|$ denotes the Euclidean distance.

For clarity, good clusterings have a large overall between-cluster distance ($SS_B$) and a small overall within-cluster distance ($SS_W$) [39]. Thus, the best partitioning of the data is reached by a high ratio of $SS_B/SS_W$, i.e. the higher value of the CH index, the better grouping of the samples is obtained.

### 2.3.3 Dunn Index

Dunn Index (DN) [16] relies on the distances among clusters and the cluster diameters. It uses the minimum pairwise distance between samples in different clusters, as the inter-cluster separation, and the maximum diameter among all clusters, as the intra-cluster compactness. This index is calculated as follows:

$$DN = \frac{\min_{1 \le i \ne j \le N} [\delta(C_i, C_j)]}{\max_{1 \le k \le N} [\Delta(C_k)]} \qquad (7)$$

In (7), $N$ represents the number of clusters, $\delta(C_i, C_j)$ represents the minimum distance between the clusters $C_i$ and $C_j$, and is described as:

$$\delta(C_i, C_j) = \min_{x \in C_i, y \in C_j} d(x, y) \qquad (8)$$

Where $d(x, y)$ denotes the distance (here euclidean) between the points $x$ and $y$. In (7), $\Delta(C_k)$ is the diameter of the cluster $C_k$, and is represented as:

$$\Delta(C_k) = \max_{x,y \in C_k} d(x, y) \qquad (9)$$

Thus, the highest DN value, the better grouping of the samples is estimated.

### 2.3.4 Bezdek index

Bezdek and colleagues proposed different indices [40]–[42], but here we will consider one of the latest indices [21], [43], which is a very effective variation of the DN index. Indeed, Bezdek recognized that the DN is very noise sensitive and proposed an improved method called generalized Dunn index (GDI), that in this article for brevity we will refer to as Bezdek index (BZ).

The variations from DN concern equation (8), which is now denoted as:

$$\delta(C_i, C_j) = \frac{1}{|C_i||C_j|} \sum_{x \in C_i, y \in C_j} d(x, y) \qquad (10)$$

where $|C|$ denots the number of points of the respective cluster and $d(x, y)$ the distance as in (8). The variation concerns as well equation (9), now denoted as:

$$\Delta(C_k) = 2\left(\frac{\sum_{x \in C_k} d(x, v_k)}{|C_k|}\right) \qquad (11)$$

where $|C_k|$ denotes the number of points of the cluster $k$, $v_k$ represents the centroid of the $k^{th}$ cluster and $d(x, v_k)$ the distance between point $x$ and centroid $v_k$. Then, $\delta(C_i, C_j)$ and $\Delta(C_k)$ are applied as in (7). Thus, the highest value of this index will represent the best estimated grouping of samples.

### 2.3.5 Davies-Bouldin index

Davies-Bouldin index (DB) [17] is calculated as follows:

$$DB = \frac{1}{N} \sum_{i=1}^{N} R_i \qquad (12)$$

In (12), $N$ denotes the number of clusters, while the $R_i$ factor is given by:

$$R_i = \max_{j \ne i} \frac{S_i + S_j}{d_{ij}} \qquad (13)$$

where the distance between the cluster centers is defined as their Euclidean distance $d_{ij} = \left\|v_i - v_j\right\|$; and $S_i$ and $S_j$ are respectively the within-cluster scatter for $i^{th}$ and j$^{th}$ clusters. $S_i$ is represented by:

$$S_i = \frac{1}{n_i} \sum_{x \in C_i} \|x - v_i\| \qquad (14)$$

where $n_i$ is the number of samples $x$ in the cluster $C_i$, $v_i$ is the center of this cluster, and $\|\cdot\|$ denotes the Euclidean distance. In this case, the smaller values of $DB$ indicate a better partitioning of the data (clusters are considered most optimally separated from each other).

In order to facilitate the comparison between the different indices of this study, since $DB$ is the only reported index where the minimum value represents the best grouping of the samples, we have inverted its value using the following formula:

$$DB^* = \frac{1}{1 + DB} \qquad (15)$$

where $DB$ is the original value returned by the index in (12). The highest $DB^*$ value, the better grouping of the samples is evaluated.

### 2.3.6 Thornton index

Thornton index (TH) [22] (also known as geometrical separability index or GSI) is defined as the fraction of a set of data points whose classification labels are the same as those of their first nearest neightbour. This index is represented as:

$$TH = \frac{1}{n}\sum_{i=1}^{n} f\left(x_i, x_i'\right) \quad (16)$$

where $x_i'$ is the first nearest neighbour of $x_i$, $n$ is the number of points, $f$ is a binary function returning either 0 or 1, depending on which class label is associated with $x_i$ and $x_i'$, and it is defined as:

$$f\left(x_i, x_i'\right) = \begin{cases} 1, if\ label\ x_i = label\ x_i' \\ 0, if\ label\ x_i \neq label\ x_i' \end{cases} \quad (17)$$

The value of TH will be closer to 1 for a set of points in which those with opposite labels exist in well-separated groups (being 1 the maximum separability value). When the groups move closer and points from opposite classes begin to geometrically overlap, the value of the index decreases. Finally, if the centroids coincide, or the points are uniformly distribuited in the space, this separability index will be close to 0.5. Thus, the highest value of this index will represent the best estimated grouping of samples.

### *2.3.7 Projection Separability Indices (PSIs)*

The Projection Separability (PS) rationale is specifically proposed to assess group separability of data samples in a geometrical space such as in the case of dimensionality reduction analyses. The main idea is to geometrically project the samples (points) of two different groups on the line that connects their centroids (Fig. 1a, b, and c). Then, considering the relative distance of the points on this line, any generic measure of separability can be applied (Fig. 1d). This procedure is repeated for all pairs of groups for which separability evaluation is desired, and then an overall estimation is computed.

On the basis of PS rationale, we propose three statistical separability measures, which we call Projection Separability Indices (PSIs). The first index, PSI-P, evaluates the separability of the points on the projection line by means of the Mann-Whitney test p-value [44]. The second index, PSI-ROC, adopts as separability measure on the projection line the Area Under the ROC-Curve (AUC-ROC) [45]. The third index, PSI-PR, uses instead the Area Under the Precision Recall Curve (AUC-PR) [46]. In general, any separability measure - such as the F-score[47], the Matthews correlation coefficient [48], and many others - can be employed, but in this study we will focus on those three measures mentioned above because they are widely used in data analysis and they are also sufficiently diverse between them to cover different types of separability estimations; The first is a ranking based statistical test, the second is a measure of trade-off between sensitivity and specificity, and the third is a measure of trade-off between precision and sensitivity (a.k.a. recall).

A general workflow of PSI is illustrated in Fig. 1. Here, an artifitial dataset was created using a web-tool (available at https://www.librec.net/datagen.html) that allows to draw the datapoints in a two-dimensional space, and get automatically the coordinates and the classes associated to each sample. In this example, two main groups were drawn: an apple (red) and a tree branch (green). First, each group centroid is identified (Fig. 1a) by calculating either the mean, median, or mode of each group (median was applied for all here presented analyses). Then, a line segment is drawn between the centroids of the two compared groups (Fig. 1b), subsequently all the points are projected onto this line using the dot product as follows:

$$proj\left(P, line(A,B)\right) = A + \frac{(AP \cdot AB)}{(AB \cdot AB)} * AB \quad (18)$$

where $P$ is the point to be projected on the line that connects the centroids $A$ and $B$ of two clusters, $AP$ and $AB$ are vectors formed by the points in question.

Once projected (Fig. 1c), the points in the line are transformed from an N-dimensional space to a 1-dimensional (1D) space, by fixing one point as reference ($p_0$) and taking the distance from it for the rest of the other points ($DP$). Then, a separability value is computed by means of any separability measure; in the specific case we applied MW p-value, AUC-ROC, and AUC-PR (Fig. 1d). In the case that more than two groups are present in the dataset, all the p-values, AUC-ROC, and AUC-PR between all the possible pair-groups are computed, and the average of each of these measures is chosen as an overall estimator of separation between the groups. Finally, the obtained values are represented as PSI-P, PSI-ROC, and PSI-PR respectively.

## 2.4 Computing platform

All algorithms were tested on MATLAB R2016b in a workstation with Windows 7 Professional with 24 GB of RAM and 6 Intel(R) Xeon(R) CPU X5660 processors with 2.80 GHz. The simulations of each dataset were executed using the High Performance Computing and Storage Complex (HRSK-II) of the Centre for Information Services and High-performance Computing (ZIH) of the Technische Universität Dresden (TU Dresden), Germany.

## 2.5 Data analysis

### *2.5.1 Normalization and algorithms' tuning*

First, in order to scale and adjust the raw values of the datasets before the analyses, the following normalizations were applied: DRS, Dividing each row by the row (samples) sum; DCS, Dividing each column by the column (features/variables) sum; LOG, Logarithmic function in base 10 of the matrix values plus 1 (because of zero values); NON, the original data without normalization were analyzed as well.

After the normalization processing, diverse DR methods were applied (they are discussed at the beginning of the Method section), obtaining respectively different separation of the samples in groups (clusters), which were afterwards assessed by means of the diverse indices. For parameter-dependent methods, such as Isomap and t-SNE, the selection of the best tuning parameters plays a crucial role in order to obtain the best group separability. In the case of Isomap, the number of the nearest neighbours (parameter $k$) modifies the construct of the proximity graph. In the case of t-SNE, we need to customize the initial dimensions used to pre-process the data by PCA ($d$), moreover, we have also to specify the perplexity ($p$) of the Gaussian kernel employed in the analysis. In order to find the best parameters setting, a tuning process was implemented that selects automatically the best combination of DR-method tuning parameters according to each index.

### 2.5.2 Statistical evaluation of separability significance

In order to asses the extent to which the values produced by the different indexes are reliable under uncertanty, for a given DR method result (in all the datasets) we frizzed the geometrical position of the samples and we reshuffled uniformely at random their group labels. This procedure was repeated 1000 times and the value of a certain index was measured at each round, generating a null model distribution composed of 1000 random values. Then, a p-value (that express the extent to which the true index value can be generated at random) was computed as the number of random values that surpass the true index value.

For each index, the comparison with the null model was characterized by: a) the mean of the 1000 null model random values; b) the standard error of the 1000 null model random values; c) the index separability significance (p-value) to quantify whether the evaluation provided by the index is significant in comparison to the null model. As we reported above - and here expressed in a different form - the separability significance represents the fraction of random reshuffling in which the index behaves better than in the true label assignement case. In other words, it measures how likely the index value will be obtained by chance. We considered significant all the p-values lower than 0.01, with respect to the null model distribution.

In the case of t-SNE and Isomap, we corrected all the p-values obtained for each tuning parameter's combination by means of Benjamini-Hochberg method [49].

### 2.5.3 Assessment and visualization of similarity between performance of different indices

We investigated the similarity between the different indices, in particular to discover the ones offering comparable results to our PSIs, regardless of the dataset or the method of embedding. To do so, we created an array that characterizes each index and has for features the index values for all the DR methods with the different data normalization, centering versions and, in the case of MCE, multiple type of distances (Euclidean distance, Spearman rank correlation distance, Pearson correlation distance). Therefore, we created a matrix that has for samples the indices and for features their values according to different DR techniques. Then, PCA was applied in order to visualize and to compare the similiarities between the separability indices arrays. Note that previous to PCA, a zscore normalization was applied to the rows (indices) in order to scale their values. This was done first separately for each dataset, and secondly by merging in a unique array the indices' results of all the datasets.

## 3 Results

The separability of the different sample groups according to the DR methods was evaluated by means of the studied indices, including the here presented PSIs, for all the datasets described in the previous section. The effectiveness of the indices is given by the correct identification of the DR method that finds the best low dimensional representation of the data, i.e. the method that shows the best group separability of the samples. The results validated by PSI-PR are presented for each dataset, in Fig. 2 (Tripartite-Swiss-Roll), Fig. 3 (Gastric mucosa microbiome), Fig. 4 (Radar signal), Fig. 5 (Image proteomics), Fig. 6 (MINST 2D), and Fig. 7 (MNIST 3D). In these figures, we report the PSI-PR value to allow the comparison between visual perception of separability and the quantification of one of the PSIs. For this measure, the closest value to one indicates the best group separability. The analougus figures obtained by the other indices are shown in supplementary information (Suppl. Fig. 1-73).

Moreover, in the same figures, for each index we report the relative separability significance calculated according to the procedure described in the methods section 2.5.2. In particular, the mean and standard error of the null model, and the separability significance (expressed as a p-value) are reported. In all the figures regarding the indices, the mean and standard error of the null model are placed in brackets next to the real value obtained by each separability index, and they represent a reference to realize how far the real estimated separation is from a null model average. On the other hand, the p-value used for evaluation of separability significance is instead placed under the value of each separability index, and it indicates how significant the reported real separability value is in comparison to a null model.

The best DR results obtained for each index are shown in Table 2 (Tripartite-Swiss-Roll), Table 3 (Gastric mucosa microbiome), Table 4 (Radar signal), Table 5 (Image proteomics), Table 6 (MNIST 2D), and Table 7 (MNIST 3D). In the tables, the order of the different DR methods was obtained by averaging the ranks (named as AVG rank) of the values provided by the indices in each case. Finally, the indices were also applied to the data in the high-dimensional space (HD) in order to see how good the DR methods are in preserving the group separability present in the original multidimensional space.

### 3.1 Tripartite-Swiss-Roll

We first present the analysis of the synthetic dataset, the Tripartite-Swiss-roll dataset that contains non-linear structures (Fig. 2a) with three clusters. In this case, all the indices placed Isomap, MCE, and t-SNE close to HD (Table 2), proving the goodness of these DR methods in preserving the original group separability present in HD. Moreover, all the indices yielded Isomap as the DR method that provided the best group separability of the data. However, in Table 2 we can see that in the case of Isomap the separability values obtained by DN, BZ, and CH are extremely high in comparison with the rest of DR methods. We might think that these indices are very accurate in terms of emphasizing the best method. Nonetheless, if we analyze carefully their performance (Suppl. Fig. 3, 5, and 6, respectively), we can clearly realize that they estimate the highest separability for Isomap when the algorithm is tuned according to a setting that provides a 2D-embedding which does not correctly preserve the instrisic structure of the Tri-partite-Swiss-Roll (Fig. 2a), situation that was as well replicated by DB* and SH (Suppl. Fig 4 and 7). Indeed, in the figures we can clearly notice that all the sample points of each respective synthetic group collapse in the same position. This explains why these indices obtained such high values: because they maximize the inter-group separation and neglect the intra-group variability, which is an important propriety to retain. Moreover, the mentioned indices found only one optimal value of tuning parameter for Isomap ($k$ = 4). In contrast, PSIs (PSI-P, PSI-ROC, and PSI-PR) and TH were capable to recognize multiple optimal $k$'s (14 options, Suppl. Fig. 9-12), where six solutions presented a perfect group separability preserving the intrinsic data structure in the first two dimensions of mappings, and a conservation of inter-group variability that is visible because of the preservation of the color gradient present in the original shape of Tripartite-Swiss-Roll (Fig. 2a). Thus, we propose the use of PSIs not only as indices for the evaluation of group separability, but as well as methods to automatically identify the best tuning parameters of paremeter-dependent algorithms such as Isomap.

Another result related with Tripartite-Swiss-Roll analysis is that there was not any index indicating PCA as best dimensionality-reduction method for this dataset. As already proven in [12], PCA is a linear transformation that cannot address these types of data nonlinearity. Therefore this result indicates that all indices were able to detect the unsuitableness of PCA in comparison to other algorthims tailored for nonlinear dimensionality reduction.

### 3.2 Gastric mucosa microbiome

This dataset provide a scenario with a small number of samples. This might represent a problem for the evaluation of the separability, because in case of low number of samples, after dimension reduction the geometrical localization of the points could be close to their random distribution.

We tested whether the DR methods could separate the three main groups: untreated dyspeptic patients without *H. pylori* infection (HPneg); untreated dyspeptic patients with H. pylori infection (HPpos); and patients treated with Proton Pump Inhibitors (PPI). In Table 3, we can appretiate that the different indices selected different DR methods that provided the best group separability of the samples, namely: t-SNE by PSI-P, DN, DB*, TH, and BZ; and MCE by PSI-PR, PSI-ROC, CH, and SH. Particularly, DN results (Suppl. Fig 19) erroneously indicate that MDSbc, MCE, PCA, and NMDS provide embedding not significantly different from the random baseline (separability significance has p-value = 1), meaning that the evaluations given by DN are not reliable for these methods. As well, for NMDS and PCA, the index BZ does not evaluate a significant separability (Suppl. Fig 21). In the case of TH, MCE, Isomap, PCA, and NMDS (methods identified by this index with significant group separability) showed projections with outlier-points belonging to the class HPpos (Suppl. Fig. 24a, c, d, and e). This could be explained by the inhability of this index to take into account how far a sample is from the rest of the clusters [50].

In general, t-SNE, Isomap, and MCE outdoes HD in sample separation in this dataset (Table 3), confirming their capability in detecting the original group separability. However, according to our PSIs, MCE was the only method able to identify a clear visual separation of the groups PPI, HPneg, and HPpos (Fig. 3, Suppl. Fig. 17-24). Moreover, these methods did not show an internal separation of PPI-treated patients according to their *H. pylori* infection. It is known in literature that PPIs cause a major perturbation in the gastric tissue microbiota of dyspeptic patients regardless of the initial pathological infection due to *H. pylori* [32]. In order to confirm our results, we repeated the analysis restricting the dataset samples only to the two PPI sub-classes (PPIneg and PPIpos). All the DR methods both linear and non-linear were not able to identify a significant group separability of these two sub-classes (Suppl. Fig. 25-37).

### 3.3 Radar signal

We analyzed as in [34], [35] the Radar signal dataset (Fig. 4) which consists of three clusters (good radar signal, bad radar signal 1, and bad radar signal 2), to see whether the separability indices are able to evaluate the results of the DR methods in presence of the crowding problem (i.e. overlap of clusters in the reduced dimensional space). We discovered that TH, SH, CH, DB*, PSI-PR, PSI-ROC, and PSI-P detected MCE as the best DR method (Table 4) which offers a group separability that according to these indices is able to outperform even the values of the same indices in HD. In the results provided by TH (Suppl. Fig. 45), SH (Suppl. Fig. 44), CH (Suppl. Fig. 43), DB* (Suppl. Fig. 41), PSI-PR (Fig. 4), PSI-ROC (Suppl. Fig. 39), and PSI-P (Suppl. Fig. 38) we can clearly notice that MCE is able to solve the crowding problem and provides a better embedding of the data into a low dimensional space as already seen in [35]. Instead, the other DR methods such as PCA, Isomap, NMDS, MDSbc, and t-SNE tend to mix the different groups (highly crowded embedding results).

In contrast, DN and BZ did not return reliable results for this dataset. In the case of BZ, this index presented values under the random baseline for Isomap, PCA, NMDS, and MDSbc. In the case of MCE, the value of this index is equal to the random baseline, however, its separabality is not significant (p-value = 0.51), meaning that this result does not provide a trustworthy group separability. Only t-SNE had a BZ value above the random baseline (Suppl. Fig. 42), however, in the figure we can clearly notice an erroneous group separability which is not significant (p-value = 1) with a predominance of the class good radar signal, which is practically distributed across all the different groups. A similar situation was shown by DN that returned values under (or equal to) the random baseline for t-SNE, NMDS, PCA, MDSbc, and MCE (Suppl. Fig. 40). For instance, in Suppl. Fig. 40a we can see that DN evaluates t-SNE result as comparable to a random distribution of the samples.

### 3.4 Image proteomics

In the case of the image proteomic dataset, we tested whether the DR methods could separate the three main groups: control patients (C), patients without neuropathic pain (M), and patients with pain (P).

In general, the indices placed t-SNE, MDSbc, Isomap, and MCE close to HD, corroborating their ability to conserve the original high dimensional group separability also in the reduced dimensional space. Specifically, all the indices placed t-SNE as the best DR method.

There is again a problem with the separability estimation provided by BZ and DN. BZ (Suppl. Fig. 50) reports values under the random baseline for MCE, MDSbc, NMDS and PCA. DN (Suppl. Fig. 48) presents the same problem regarding the evaluation of PCA, NMDS, and MDSbc. Moreover, we noticed that in the embedding selected by CH (Suppl. Fig. 51), all the DR methods display a fuzzy structure of the groups related to the classes C and P (both groups are mixed). This situation is shared by PCA and NMDS for all the different indices (Fig. 5, Suppl. Fig. 46-53). Interestingly, t-SNE presented a perfect group separability according to all PSIs (Fig. 5, Suppl. Fig. 46-47), DB* (Suppl. Fig. 49), and TH (Suppl. Fig. 53). In the case of the t-SNE projection selected by SH (Suppl. Fig. 52), all groups are splitted. Moreover, TH and PSIs identified multiple tunning parameters for t-SNE (where the other indices identified only one). TH found six possible solutions of t-SNE tunning parameters (Suppl. Fig. 57) which allow its maximization, while PSIs only detected three possible solutions of t-SNE tunning parameters (Suppl. Fig. 54-56), which maximize separability. If we compare these results, we can notice that in half of the solutions returned by TH, either the class P or M is highly splitted in multiple sub-groups (Suppl. Fig 57a-c). In contrast, the solutions identified by PSIs (Suppl. Fig 54-56), the three groups (C, M, and P) are mostly perfectly segregated. This demostrates the utility of our PSIs to identify automatically the best combinatorial tuning of parameters for a parameter-dependent algorithm such as t-SNE.

### 3.5 MNIST

One of the limitations of the previous tests is that we considered datasets composed by not more than 4 groups, therefore the embedding in 2D space was enough to represent the variability between this limited number of groups. In order to overcome this restriction, in this section we investigate the performance of the separability indices in a more diversified scenario, which involves a dataset composed of 10 groups. Hence, here, we will need to evaluate the performance of embedding not only in 2D but also in 3D space. To this aim, we analyzed a subset of the MNIST dataset in which, from the original 60000 training images, 300 samples were randomly selected (30 samples for each number from zero to nine) in order to create 10 groups. Each group containes 30 samples of images representing the same number with different handwrite variations. Then, two analyses, one in a two-dimensional space (2D) and another one in a three-dimensional space (3D) of embedding were carried out in order to compare the group separability given by the DR methods in a different number of dimensions. Interestingly, all the indices placed t-SNE as best DR method in both 2D and 3D spaces (Table 6 and Table 7), in fact, in the tables we can notice that this technique is the only one outperforming HD. t-SNE is able to separate the digit classes but, as for the other DR methods, some numbers that are difficult to identify (distorded digits) may be clustered in the wrong group (Fig. 6-7, Suppl. Fig. 58-73).

Again, BZ and DN presented evaluations under the random baseline. BZ presented this problem for MDSbc, NMDS, PCA, MCE, and Isomap in both low-dimensional spaces (Suppl. Fig. 62 and Suppl. Fig. 70). DN presented also the same problem for PCA, MCE, MDSbc, and NMDS in both dimensional spaces (Suppl. Fig. 60 and Suppl. Fig. 68).

MNIST dataset has the peculiarity that does not present a hierarchical organization of the samples. Indeed, the graphical shape of numbers is not organized in a hierarchy. Hence, MCE should be evaluated with low performance for this dataset, because it is not a manifold embedding but a hierarchical embedding method (for more information see [35]). Interestingly, PSI-PR (Fig. 6 and Fig. 7) and PSI-ROC (Suppl. Fig. 59 and Suppl. Fig. 67) detected this situation, evaluating MCE as the last DR technique in both dimensional spaces (2D and 3D).

Moreover, looking at the results returned by PSIs (Table 6 and 7) we notice that in presence of hundreds of samples and a wide number of groups, a 3D visualization of the DR methods works – as expected - a bit better than a 2D representation. Indeed, the representation of group separability is slightly more evident in a three-dimensional space, in which, the different groups of numbers look more separated and less overlapped than in a two-dimensional space. As anticipated in the introduction this comparison between separability performance in different dimensions is less straitforward if we employ methods such as CVIs that do not have a bounded range of evaluation, therefore the same value in two different dimensional space of different size does not necessarily indicate a similar level of separation.

### 3.7 Similarities across different indices

A final comparison was carried out in order to analyze the similarities between the different indices in each dataset (Suppl. Fig. 74), by means of PCA. It might seem outlandish, we indeed adopt dimensionality reduction to evaluate the similarity between techniques that are designed to quantify the separability obtained by dimension reduction methods. In Suppl. Fig. 74 and Fig. 8, we have created a triangle whose vertices are the PS metrics (PSI-P, PSI-ROC, and PSI-PR), named as PSI-triangle, with the aim to visually spot what indices are comparable to our PSIs (the ones inside the PSI-triangle).

In the case of Tripartite-Swiss-Roll dataset, TH is close to the PSI-triangle, meaning that this index gave similar validation results to PSI. In fact, these indices do not maximize the internal geometrical connectivity as confirmed for this dataset. Moreover, in the case of MNIST (3D) dataset, BZ is also close to the PSI-triangle. In fact, if we compare Fig. 7 and Supp. Fig. 70, we can notice that there are similar dimensionality reduction structures selected by these two separability measures. However, our PSIs provided in contrast to BZ a significant separability in comparison to a random permutation of the labels.

Finally, to see the overall trend across all the datasets, the analysis was repeated by merging the indices results of all the datasets (Fig. 8). It confirms again the coherence of the different statistical PS measures which were specifically design for the validation of group separability returned by DR techniques.

In Fig. 8, if we project all indices on the first dimension (the one with the highest data variability), we can notice that TH, SH, and DB* are the closest indices in relation to the PSIs. SH and DB have been catalogated as the best cluster validity indexes in several studies [14], [51]–[53]. Given the above, their proximity with our PSIs might confirm that our measures performe well. However, they present some limitations that our PSIs do not. DB is not designed to accommodate overlapping clusters, and on the other hand, SH is only able to identify the first best choice and therefore should not be applied to datasets with sub-clusters [54]. The exclusion of these limitations by our measures is confirmed by their proximity in Fig. 8 with TH, due to the fact that our PSIs and TH do not maximize the internal geometrical connectivity of the groups. Thus, PSIs and TH do not force the samples of each class to collapse in a unique point: one unique point for each cluster would be a representation that neglects the internal differences between the samples of a cluster. Therefore, the adoption of PSIs reduce the risk to penalize the visual evidence of subclasses after dimensionality reduction.

## 4 Discussion

There is not a universal way to map a given dataset from the high-dimensional space into a reduce number of dimensions by perfectly preserving all the proprieties of the original structure. Thus, the embedding performance of different DR methods are often the result of 'computational trade-off', which sacrifices some properties in order to preserve others. For example, PCA tries to preserve linear structures, classical MDS tries to preserve global geometry, t-SNE tries to preserve local proprieties and local density of the data , MCE tries to preserve hierarchy, and so on [55]. In this context, we propose PS as a novel rationale in order to design separability measures tailored to compare the performance of different DR methods, or the performance of a single DR method across different dimensions of embedding. We also propose a new class of indices named PSIs for evaluation of the separability performance. In this moment this class includes three indices called PSI-P, PSI-ROC, PSI-PR that are based on three accepted measure of statistical separability widely adopted in machine learning. However, in the future other indices could be proposed by simply implementing any other statistical measure (for instance the Pearson correlation coefficient) according to the PS rational.

We compared the effectiveness of the three proposed PSIs versus the state-of-the-art CVIs in evaluating different DR methods across multiple datasets. DN and BZ returned some untrustworthy group separability results whose values were equal or even lower than the random baseline (with a separability significance (p-value) > 0.01 in Gastric mucosa microbiome, Radar signal, Image proteomics, and MNIST datasets). In contrast to these indices, PSIs obtained higher evaluation values in comparison to a random permutation of the labels for all the cases. This is an evidence that PSIs can detect, when present, better group separability than DN and BZ.

Another issue emerged in the analysis of Tripartite-Swiss-Roll dataset, in which all the indices with the exception of TH and our PSIs returned high value in the presence of collapsed groups. In fact, TH and PSI do not maximize the internal geometrical connectivity of the groups as they do not force the samples of each class to collapse in one point (each group having a different point). This reduces the potential risk to cancel the visual evidence of subclasses after dimensionality reduction. In particular, DN, BZ, SH, CH, and DB* detected a unique setting of the Isomap tuning parameter that offered the best separability. Instead, TH and PSIs were capable to identify multiple settings of the Isomap tuning parameter that offered a comparable best separability (Suppl. Fig 9-12), and this was in agreement with the visual perception of the results. The same results appeared also for t-SNE (Suppl. Fig 13-16). Notably, some tuning parameter settings of Isomap and t-SNE selected by TH and PSIs exhibited finer representations of the original non-linear structure by conserving the initial color gradient and a clear visualization of the groups. However, TH returned some equivalent solutions of best tuning for t-SNE with a questiobale segregation of the groups (Suppl. Fig. 16a-h). This appears as well in the analysis of the image proteomic dataset where TH selected more equivalent best tuning options than PSIs, but PSIs selected embeddings with a more precise group separability, therefore PSIs offered evidences to be more precise than TH, at least in these datasets. Hence, our novel group separability indices can also be employed to enhance the tuning of dimensionality reduction algorithms. However, we should take into account that in some cases there might be more information associated with subgroups in the data that can be lost by forcing the group separability with different tuning parameters. Therefore, parameter-free methods for non-linear DR such as MCE can help to identify hidden patterns from the original data, reducing the risk to neglect unknown sub-group separations because of tuning parameter bias. In this case, PSIs can still be used for the evaluation of parameter-free algorithms - because can help to detect in which dimensions the separability of groups emerges. Traditional CVIs are less reliable to compare results accross different dimensions as we explained in the results section above.

Finally, we analyzed the simililarities between the indices: initially separately for each dataset (Suppl. Fig. 74), and then collectively for all datasets together (Fig. 8). These analyses revealed that DB*, SH, and TH are closer to our PSIs than other indices. Moreover, TH and PSIs overcome some limitations of DB* and SH, explained by the fact that both PSIs and TH do not force the intra cluster connectivity, allowing the selection of embedding representations that retain intra cluster variability, in contrast with the rest of the indices.

To sum up, in the analyses of high-dimensional datasets and their embedding, PSIs provided a good performance in terms of separability evaluation, supporting the idea that statistical measures based on the projection separability rationale can be complementary and often can work better than traditional cluster validity indices [56] based on mere geometrical rationale such as density or continuity.

## Acknowledgment

C.D. is funded by the Research Grants – Doctoral Programs in Germany (DAAD), Promotion program Nr: 57299294. A.A and S.C have equally contributed to this manuscript. C.V.C is the corresponding author.

## References

[1] M. Belkin and P. Niyogi, "Laplacian Eigenmaps for Dimensionality Reduction and Data Representation," *Neural Comput.*, vol. 15, no. 6, pp. 1373–1396, 2003.

[2] C. V. Cannistraci, T. Ravasi, F. M. Montevecchi, T. Ideker, and M. Alessio, "Nonlinear dimension reduction and clustering by minimum curvilinearity unfold neuropathic pain and tissue embryological classes," *Bioinformatics*, vol. 26, no. 18, 2010.

[3] A. Muscoloni, J. M. Thomas, S. Ciucci, G. Bianconi, and C. V. Cannistraci, "Machine learning meets complex networks via coalescent embedding in the hyperbolic space," *Nat. Commun.*, vol. 8, no. 1, p. 1615, Dec. 2017.

[4] S. Ciucci *et al.*, "Enlightening discriminative network functional modules behind Principal Component Analysis separation in differential-omic science studies," *Sci. Rep.*, vol. 7, pp. 408–421, 2017.

[5] Shuicheng Yan, Dong Xu, Benyu Zhang, and Hong-Jiang Zhang, "Graph Embedding: A General Framework for Dimensionality Reduction," in *2005 IEEE Computer Society Conference on Computer Vision and Pattern Recognition (CVPR'05)*, vol. 2, pp. 830–837.

[6] G. Alanis-Lobato, C. V. Cannistraci, A. Eriksson, A. Manica, and T. Ravasi, "Highlighting nonlinear patterns in population genetics datasets," *Sci. Rep.*, vol. 5, 2015.

[7] K. Pearson, "On lines and planes of closest fit to systems of points in space," *Philos. Mag. Ser. 6*, vol. 2, no. 11, pp. 559–572, 1901.

[8] H. Hotelling, "Analysis of a complex of statistical variables into principal components," *J. Educ. Psychol.*, 1933.

[9] W. S. Torgerson, "Multidimensional scaling: I. Theory and method," *Psychometrika*, vol. 17, no. 4, pp. 401–419, Dec. 1952.

[10] K. Bunte, S. Haase, M. Biehl, and T. Villmann, "Stochastic neighbor embedding (SNE) for dimension reduction and visualization using arbitrary divergences," *Neurocomputing*, vol. 90, pp. 23–45, Aug. 2012.

[11] L. van der Maaten and G. Hinton, "Visualizing Data using t-SNE," *J. Mach. Learn. Res.*, vol. 9, no. Nov, pp. 2579–2605, 2008.

[12] J. B. Tenenbaum, V. de Silva, and J. C. Langford, "A global geometric framework for nonlinear dimensionality reduction.," *Science*, vol. 290, no. 5500, pp. 2319–23, Dec. 2000.

[13] A. K. Jain, M. N. Murty, and P. J. Flynn, "Data clustering: a review," *ACM Comput. Surv.*, vol. 31, no. 3, pp. 264–323, Sep. 1999.

[14] O. Arbelaitz, I. Gurrutxaga, J. Muguerza, J. M. Pérez, and I. Perona, "An extensive comparative study of cluster validity indices," *Pattern Recognit.*, vol. 46, no. 1, pp. 243–256, Jan. 2013.

[15] K. Kryszczuk and P. Hurley, "Estimation of the Number of Clusters Using Multiple Clustering Validity Indices," Springer, Berlin, Heidelberg, 2010, pp. 114–123.

[16] J. C. Dunn, "A fuzzy relative of the ISODATA process and its use in detecting compact well-separated clusters," *J. Cybern.*, vol. 3, no. 3, pp. 32–57, 1973.

[17] D. L. Davies and D. W. Bouldin, "A Cluster Separation Measure," *IEEE Trans. Pattern Anal. Mach. Intell.*, vol. PAMI-1, no. 2, pp. 224–227, 1979.

[18] T. Calinski and J. Harabasz, "A dendrite method for cluster analysis," *Commun. Stat.*, vol. 3, no. 1, pp. 1–27, 1974.

[19] P. J. Rousseeuw, "Silhouettes: A graphical aid to the interpretation and validation of cluster analysis," *J. Comput. Appl. Math.*, vol. 20, pp. 53–65, Nov. 1987.

[20] J. C. Bezdek†, "Cluster Validity with Fuzzy Sets," *J. Cybern.*, vol. 3, no. 3, pp. 58–73, Jan. 1973.

[21] B. Stein, S. M. zu Eissen, Gf. WiBbrock, and M. H. Hanza, "On cluster validity and the information need of users," no. September, pp. 216–221, 2003.

[22] C. Thornton, "Separability is a Learner's Best Friend," Springer, London, 1998, pp. 40–46.

[23] D. G. Calò, "On a transvariation based measure of group separability," *J. Classif.*, vol. 23, no. 1, pp. 143–167, 2006.

[24] D. J. Hand, *Construction and assessment of classification rules*. Wiley, 1997.

[25] C. Domeniconi, D. Gunopulos, S. Ma, B. Yan, M. Al-Razgan, and D. Papadopoulos, "Locally adaptive metrics for clustering high dimensional data," *Data Min. Knowl. Discov.*, vol. 14, no. 1, pp. 63–97, Feb. 2007.

[26] C. Domeniconi, D. Papadopoulos, D. Gunopulos, and S. Ma, "Subspace Clustering of High Dimensional Data," in *Proceedings of the 2004 SIAM International Conference on Data Mining*, 2004, pp. 517–521.

[27] M. Verleysen and D. François, "The Curse of Dimensionality in Data Mining and Time Series Prediction," Springer, Berlin, Heidelberg, 2005, pp. 758–770.

[28] C. O. S. Sorzano, J. Vargas, and A. P. Montano, "A survey of dimensionality reduction techniques," pp. 1–35, 2014.

[29] R. N. Shepard, "The analysis of proximities: Multidimensional scaling with an unknown distance function. I.," *Psychometrika*, vol. 27, no. 2, pp. 125–140, Jun. 1962.

[30] J. B. Kruskal, "Multidimensional scaling by optimizing goodness of fit to a nonmetric hypothesis," *Psychometrika*, vol. 29, no. 1, pp. 1–27, Mar. 1964.

[31] E. W. Beals, "Bray-curtis ordination: An effective strategy for analysis of multivariate ecological data," in *Advances in Ecological Research*, vol. 14, no. C, 1984, pp. 1–55.

[32] F. Paroni Sterbini *et al.*, "Effects of Proton Pump Inhibitors on the Gastric Mucosa-Associated Microbiota in Dyspeptic Patients.," *Appl. Environ. Microbiol.*, vol. 82, no. 22, pp. 6633–6644, Nov. 2016.

[33] V. G. Sigillito, S. P. Wing, L. V. Hutton, and K. B. Baker, "Classification of radar returns from the ionosphere using neural networks," *Johns Hopkins APL Tech. Dig. (Applied Phys. Lab.*, vol. 10, no. 3, pp. 262–266, 1989.

[34] A. D. Shieh, T. B. Hashimoto, and E. M. Airoldi, "Tree preserving embedding.," *Proc. Natl. Acad. Sci. U. S. A.*, vol. 108, no. 41, pp. 16916–21, Oct. 2011.

[35] C. V. Cannistraci, G. Alanis-Lobato, and T. Ravasi, "Minimum curvilinearity to enhance topological prediction of protein interactions by network embedding," in *Bioinformatics*, 2013, vol. 29, no. 13.

[36] A. Conti *et al.*, "Pigment epithelium-derived factor is differentially expressed in peripheral neuropathies," *Proteomics*, vol. 5, no. 17, pp. 4558–4567, Sep. 2005.

[37] A. Conti *et al.*, "Differential expression of ceruloplasmin isoforms in the cerebrospinal fluid of amyotrophic lateral sclerosis patients," *PROTEOMICS - Clin. Appl.*, vol. 2, no. 12, pp. 1628–1637, Dec. 2008.

[38] Y. Lecun, L. Bottou, Y. Bengio, and P. Haffner, "Gradient-based learning applied to document recognition," *Proc. IEEE*, vol. 86, no. 11, pp. 2278–2324, 1998.

[39] N. Tahiri, M. Willems, and V. Makarenkov, "A new fast method for inferring multiple consensus trees using k-medoids.," *BMC Evol. Biol.*, vol. 18, no. 1, p. 48, 2018.

[40] Bezdek, "Cluster Validity with Fuzzy Sets," *J. Cybern.*, vol. 3,

no. 3, pp. 58–73, 1973.

[41] J. C. Bezdek, *Pattern Recognition with Fuzzy Objective Function Algorithms*. Boston, MA: Springer US, 1981.

[42] D. T. Anderson, J. C. Bezdek, M. Popescu, and J. M. Keller, "Comparing Fuzzy, Probabilistic, and Possibilistic Partitions," *IEEE Trans. Fuzzy Syst.*, vol. 18, no. 5, pp. 906–918, Oct. 2010.

[43] J. C. Bezdek, W. Q. Li, Y. Attikiouzel, and M. Windham, "A geometric approach to cluster validity for normal mixtures," *Soft Comput. - A Fusion Found. Methodol. Appl.*, vol. 1, no. 4, pp. 166–179, 1997.

[44] H. B. Mann and D. R. Whitney, "On a Test of Whether one of Two Random Variables is Stochastically Larger than the Other," *Ann. Math. Stat.*, vol. 18, no. 1, pp. 50–60, 1947.

[45] J. S. Hanley and B. J. McNeil, "The Meaning and Use of the Area under a Receiver Operating Characteristic (ROC) Curve," *Radiology*, vol. 143, no. 1, pp. 29–36, 1982.

[46] V. Raghavan, P. Bollmann, and G. S. Jung, "A critical investigation of recall and precision as measures of retrieval system performance," *ACM Trans. Inf. Syst.*, vol. 7, no. 3, pp. 205–229, 1989.

[47] N. Chinchor, D. Ph, C. P. Drive, and S. Diego, "MUC-4 EVALUATION METRICS," pp. 22–29.

[48] B. W. Matthews, "Comparison of the predicted and observed secondary structure of T4 phage lysozyme," *BBA - Protein Struct.*, vol. 405, no. 2, pp. 442–451, 1975.

[49] Y. Benjamini and Y. Hochberg, "Controlling the false discovery rate: a practical and powerful approach to multiple testing," *Journal of the Royal Statistical Society*, vol. 57, no. 1. pp. 289–300, 1995.

[50] L. Mthembu and T. Marwala, "A note on the separability index," Dec. 2008.

[51] S. Chaimontree, K. Atkinson, and F. Coenen, "Best Clustering Configuration Metrics: Towards Multiagent Based Clustering," in *Advanced Data Mining and Applications*, 2010, pp. 48–59.

[52] S. Petrovic, "A Comparison Between the Silhouette Index and the Davies-Bouldin Index in Labelling IDS Clusters," *11th Nord. Work. Secur. IT-systems*, pp. 53–64, 2006.

[53] M. Kim and R. S. Ramakrishna, "New indices for cluster validity assessment," *Pattern Recognit. Lett.*, vol. 26, no. 15, pp. 2353–2363, 2005.

[54] S. Saitta, B. Raphael, and I. F. C. Smith, "A Bounded Index for Cluster Validity," in *Machine Learning and Data Mining in Pattern Recognition*, 2007, pp. 174–187.

[55] L. H. Nguyen and S. Holmes, "Ten quick tips for effective dimensionality reduction," *PLoS Comput. Biol.*, vol. 15, no. 6, p. e1006907, 2019.

[56] M. Halkidi, Y. Batistakis, and M. Vazirgiannis, "Cluster validity methods," *ACM SIGMOD Rec.*, vol. 31, no. 2, p. 40, Jun. 2002.

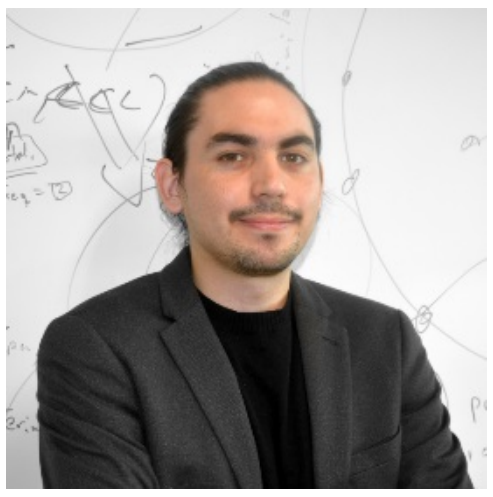

**Aldo Acevedo** received the Engineer diploma degree in Bioinformatics from the University of Talca, Chile, in 2015. He worked as full-stack developer in the Social Responsibility Department at the University of Talca from 2015 until 2016. He is currently a computer science Ph.D. student in the Biomedical Cybernetics Group led by Dr. Carlo Vittorio Cannistraci at the Biotechnology Center of the Technische Universität Dresden, Germany. His research interests include software engineering, artificial intelligence, and bioinformatics.

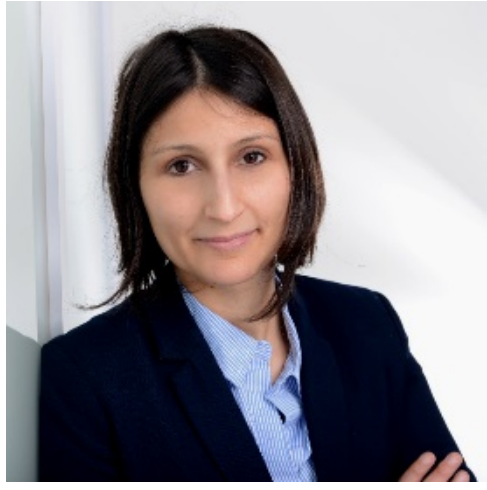

**Sara Ciucci** received the B.Sc. degree in Mathematics from the University of Padova, Italy, in 2010 and the M.Sc. degree in Mathematics from the University of Trento, Italy in 2014. She obtained her Ph.D. degree in Physics from Technische Universität Dresden, Germany, in 2018 under the supervision of Dr. Carlo Vittorio Cannistraci in the Biomedical Cybernetics Group, where she remained as postdoctoral researcher till March 2019. Her research interests include machine learning, network science, and systems biomedicine.

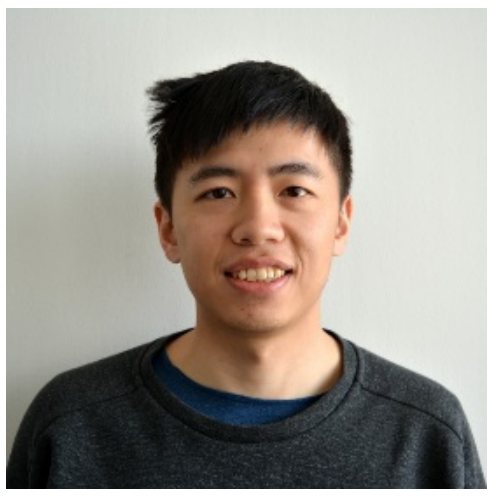

**Ming-Ju Kuo** received the B.Sc. from National Taiwan University, Taiwan, in 2015. In 2019, he obtained the Master in molecular bioengineering from the Biotechnology Center of the Technische Universität Dresden (Germany), under the supervision of Dr. Carlo Vittorio Cannistraci in the Biomedical Cybernetics Group, where he is currently scientific assistant. His interests are molecular biology and computational biology.

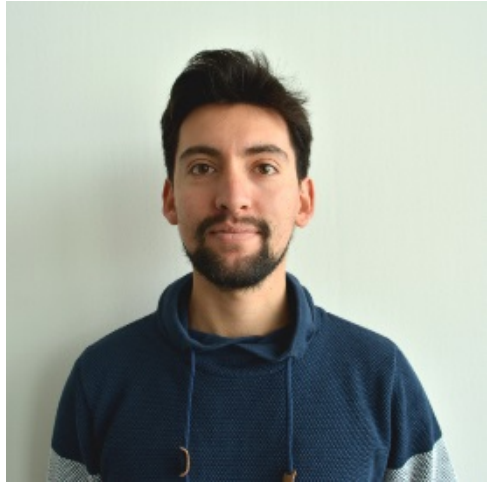

**Claudio Durán** received the Engineer diploma degree in Bioinformatics from the University of Talca, Chile, in 2016. He is currently a computer science Ph.D. student in the Biomedical Cybernetics Group led by Dr. Carlo Vittorio Cannistraci at the Biotechnology Center of the Technische Universität Dresden, Germany. His research interests include machine learning, network science, and systems biomedicine.

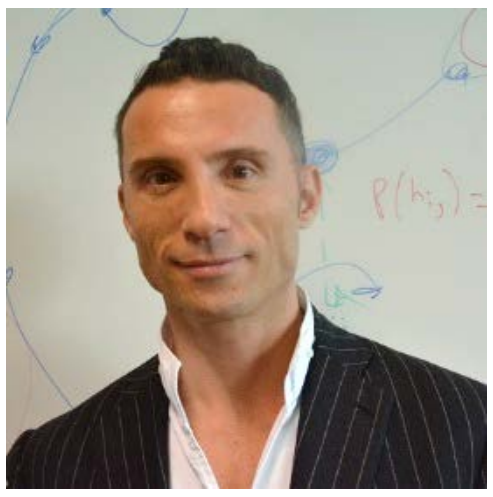

**Carlo Vittorio Cannistraci** is a theoretical engineer and was born in Milazzo, Sicily, Italy, in 1976. He received the M.S. degree in Biomedical Engineering from the Polytechnic of Milano, Italy, in 2005 and the Ph.D. degree in Biomedical Engineering from the Inter-polytechnic School of Doctorate, Italy, in 2010. From 2009 to 2010, he was visiting scholar in the Integrative Systems Biology lab of Dr. Trey Ideker at the University California San Diego (UCSD), USA. From 2010 to 2013, he was postdoc and then research scientist in machine intelligence and complex network science for personalized biomedicine at the King Abdullah University of Science and Technology (KAUST), Saudi Arabia. Since 2014, he has been Independent Group Leader and Head of the Biomedical Cybernetics lab at the Biotechnological Center (BIOTEC) of the TU-Dresden, Germany. He is also affiliated with the MPI Center for Systems Biology Dresden and with the Tsinghua Laboratory of Brain and Intelligence (China). He is author of three book chapters and more than 40 articles. His research interests include subjects at the interface between physics of complex systems, complex networks and machine learning, with particular interest for applications in biomedicine and neuroscience. Dr. Cannistraci is member of the Network Science Society, member of the International Society in Computational Biology, member of the American Heart Association, member of the Functional Annotation of the Mammalian Genome Consortium. He is Editor for the mathematical physics board of the journal Scientific Reports (edited by Nature) and of PLOS ONE. *Nature Biotechnology* selected his article (*Cell* 2010) on machine learning in developmental biology to be nominated in the list of 2010 notable breakthroughs in computational biology. *Circulation Research* featured his work (*Circulation Research* 2012) on leveraging a cardiovascular systems biology strategy to predict future outcomes in heart attacks, commenting: “a space-aged evaluation using computational biology”. TU Dresden honoured Dr. Cannistraci of the *Young Investigator Award 2016 in Physics* for his recent work on the local-community-paradigm theory and link prediction in monopartite and bipartite complex networks. In 2017, Springer-Nature scientific blog highlighted with an interview to Dr. Cannistraci his recent study on “How the brain handles pain through the lens of network science”. In 2018, the American Heart Association covered on its website Dr. Cannistraci’s chronobiology discovery on how the sunshine affects the risk and time onset of heart attack. In 2018, Nature Communications featured Carlo’s article entitled “Machine learning meets complex networks via coalescent embedding in the hyperbolic space” in the selected interdisciplinary collection of recent research on complex systems. In 2019, Dr. Cannistraci was awarded of the Shanghai 1000 plan and the Zhou Yahui Chair professorship at Tsinghua University.

TABLE 1
DATASET DETAILS

| Dataset | Samples | Features/Variables | Classes |
|---|---|---|---|
| **Tripartite-Swiss-Roll** | 723 | 3 | 3 |
| **Gastric mucosa microbiome** | 24 | 187 | 3 |
| **Radar signal** | 350 | 34 | 3 |
| **Image proteomics** | 42 | 1947 | 3 |
| **MNIST** | 300 | 784 | 10 |

*All the studied datasets are high dimensional. Hence, different dimensional-reduction (DR) methods were applied in order to embed each dataset into the first two dimentions (2D) or three dimensions (3D) of mapping.*

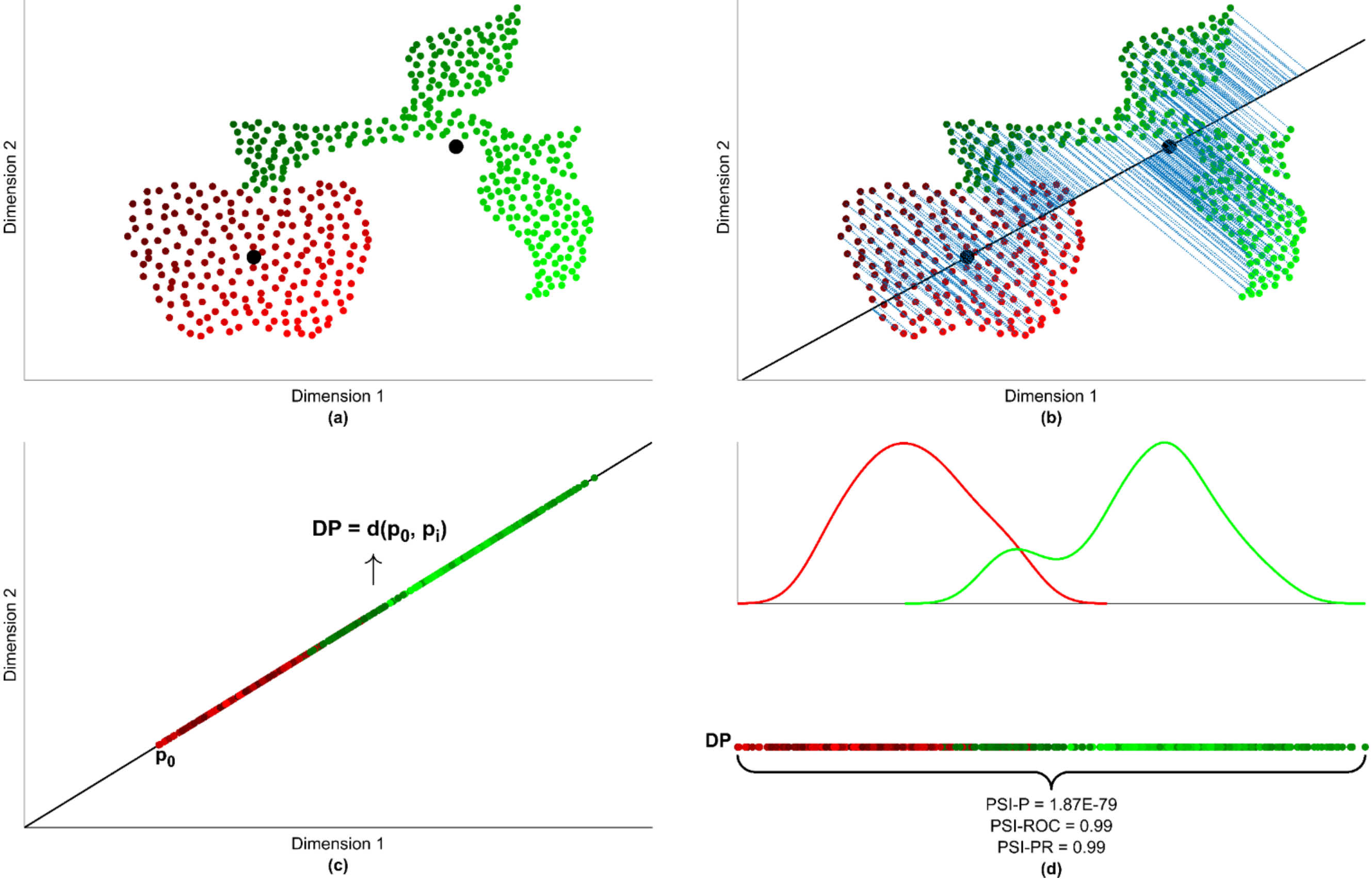

**Fig. 1. Projection Separability Index (PSI) workflow.**
In this example, we can visually identify two clusters as inputs for PSI: the apple (regular cluster) in red and the tree branch (irregular cluster) in green. The algorithm starts determining the centroids (black points) of the clusters (a), then it projects the sample points on the line that connects the cluster centroids (b). Once projected, the sample points are transformed from N-dimensional space to 1D space (the line), by fixing one extreme point as reference ($p_0$) and taking the distance from it for the rest of the other points ($DP = d(p_0, p_1)$) (c). Finally, the projection separability index is calculated computing Mann-Whitney test p-value (PSI-P), Area Under the ROC-Curve (PSI-ROC), and Area Under the Precision Recall Curve (PSI-PR) from the projected positions of the points on the line.

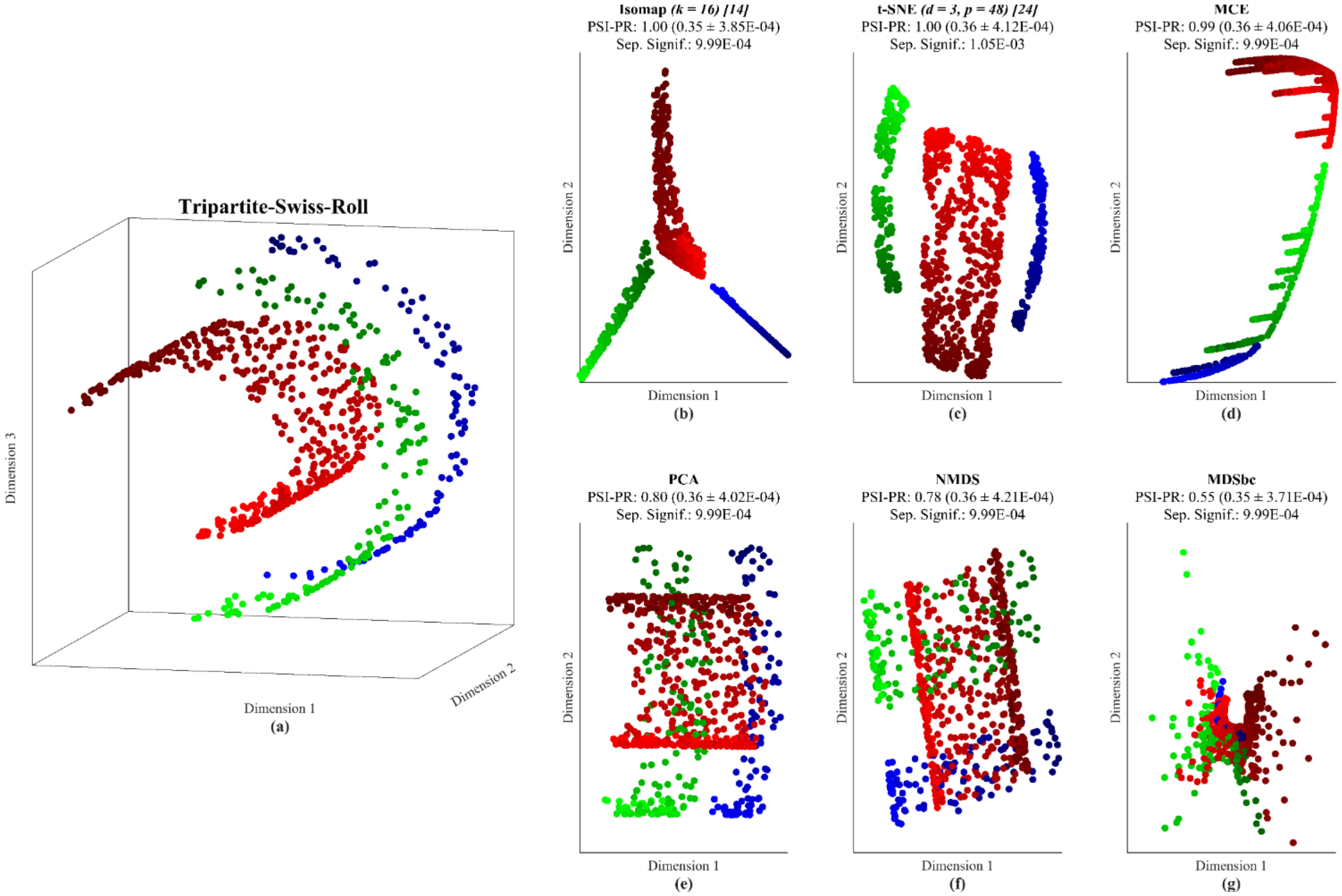

**Fig. 2. PSI-PR of the DR methods applied to Tripartite-Swiss-Roll dataset.**
In (a) it is displayed the Tripartite-Swiss-Roll dataset in the original 3D-space. The three different colours (red, blue, and green) represent the three partitions of the Swiss-roll manifold while the color gradient serves as a reference to see how good the DR methods are in preserving the original non-linear structure. The DR methods were sorted from the best (top left) to the worst (bottom right) in (b)-(g) according to the PSI-PR validation value. In (b) and (c), the optimal tuning parameter is specified in parentheses next to the title of each DR method, moreover, the digit in brackets represents the number of possible solutions with the same validation value. Furthermore, next to the value returned by this index, it is reported in parentheses the random baseline, which is the mean and the standard error of the PSI-PRs computed by randomly re-shuffling the class labels 1000 times. Finally, we added a p-value which indicates the separability significance of each index in comparison to a respective null model computed by random re-shuffling.

### TABLE 2
### VALUES OF THE DIFFERENT INDICES OBTAINED FOR TRIPARTITE-SWISS-ROLL DATASET

| Method | PSI-P | PSI-ROC | PSI-PR | DN | DB* | BZ | CH | SH | TH | AVG Rank |
|---|---|---|---|---|---|---|---|---|---|---|
| **Isomap** | 3.69E-41 | 1.00 | 1.00 | 17533.66 | 1.00 | 45464.34 | 1150067129342.85 | 1.00 | 1.00 | 1.00 |
| **t-SNE** | 3.69E-41 | 1.00 | 1.00 | 0.59 | 0.70 | 1.69 | 1214.60 | 0.86 | 1.00 | 1.67 |
| **MCE** | 5.25E-37 | 0.99 | 0.99 | 0.03 | 0.61 | 0.90 | 3421.58 | 0.63 | 1.00 | 2.56 |
| **HD** | 6.67E-41 | 0.93 | 0.87 | 0.20 | 0.35 | 0.79 | 100.28 | 0.21 | 1.00 | 3.22 |
| **NMDS** | 1.20E-23 | 0.88 | 0.78 | 1.28E-03 | 0.35 | 0.79 | 91.86 | 0.14 | 0.82 | 4.56 |
| **PCA** | 1.20E-23 | 0.88 | 0.80 | 1.28E-03 | 0.35 | 0.79 | 91.86 | 0.14 | 0.85 | 4.78 |
| **MDSbc** | 2.14E-05 | 0.74 | 0.55 | 2.33E-04 | 0.21 | 0.46 | 42.55 | -0.10 | 0.80 | 6.11 |

*This table presents the best DR validation values of the different indices for the Tripartite-Swiss-Roll dataset. In the table, the high-dimensional (HD) space was included in order to show the performance of the indices in the original high-dimensional space. Moreover, an average of rank performances (denoted as AVG Rank) of the indices values was calculated, where the lowest AVG Rank value represents the algorithm evaluated as the best DR method (as the one that offers the best group separability of the samples) by a higher number of indices.*

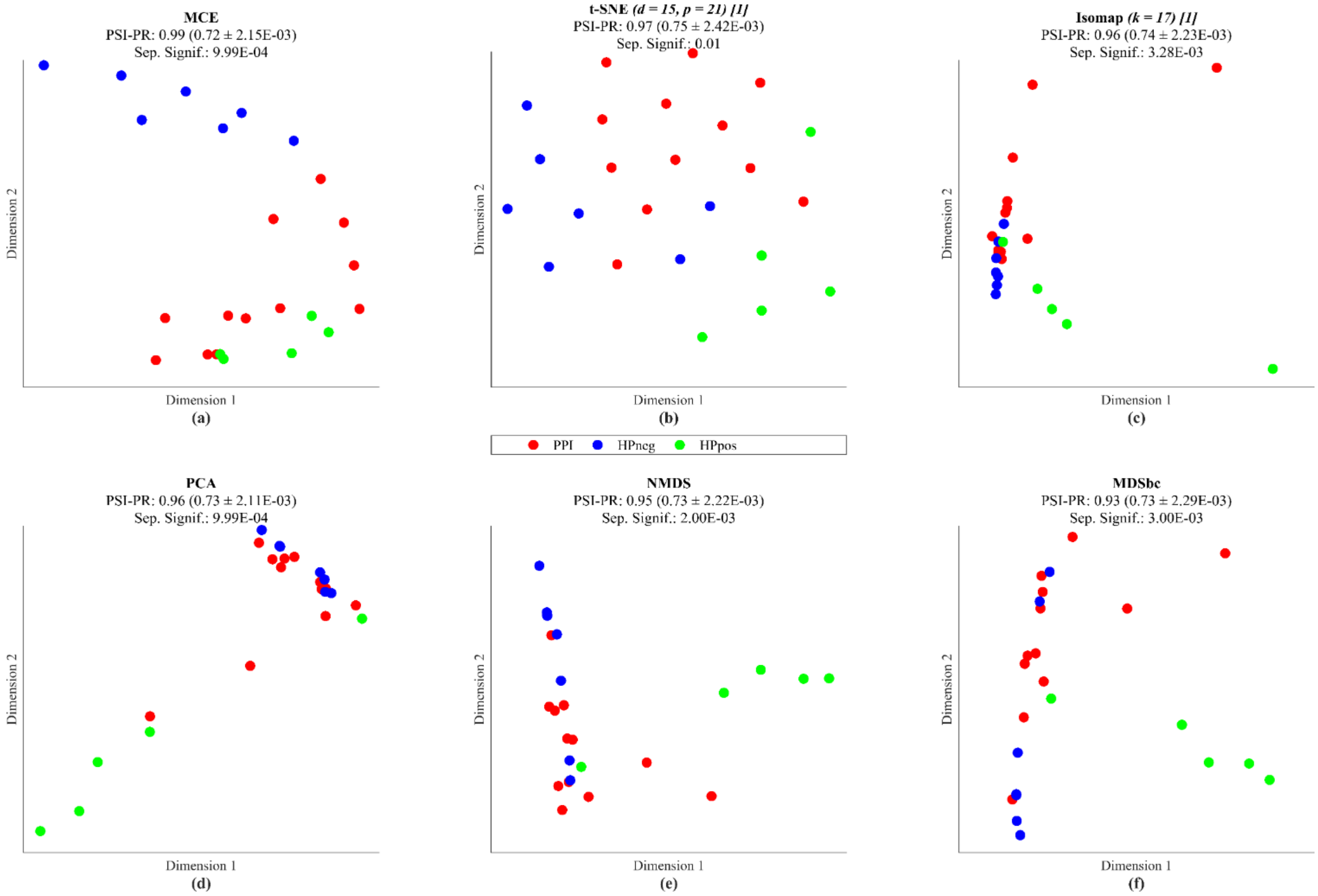

**Fig. 3. PSI-PR of the DR methods applied to Gastric mucosa microbiome dataset.**
The DR methods were sorted from the best (top left) to the worst (bottom right) in (a)-(f) according to the PSI-PR value. In (b) and (c), the optimal tuning parameter is specified in parentheses next to the title of each DR method, moreover, the digit in brackets represents the number of possible solutions with the same validation value. Furthermore, next to the value returned by this index, it is reported in parentheses the random baseline, which is the mean and the standard error of the PSI-PRs computed by randomly re-shuffling the class labels 1000 times. Finally, we added a p-value which indicates the separability significance of each index in comparison to a respective null model computed by random re-shuffling.

TABLE 3

VALUES OF THE DIFFERENT INDICES OBTAINED FOR GASTRIC MUCOSA MICROBIOME DATASET

| Method | PSIP | PSIROC | PSIPR | DN | DB* | BZ | CH | SH | TH | AVG Rank |
|---|---|---|---|---|---|---|---|---|---|---|
| **t-SNE** | 2.52E-03 | 0.96 | 0.97 | 0.26 | 0.48 | 0.94 | 18.91 | 0.41 | 0.79 | 1.56 |
| **MCE** | 3.98E-03 | 0.97 | 0.99 | 0.04 | 0.47 | 0.77 | 38.81 | 0.42 | 0.79 | 1.89 |
| **Isomap** | 0.01 | 0.93 | 0.96 | 0.09 | 0.45 | 0.85 | 15.86 | 0.32 | 0.75 | 3.33 |
| **HD** | 0.01 | 0.94 | 0.97 | 0.32 | 0.38 | 0.72 | 9.96 | 0.20 | 0.63 | 4.33 |
| **PCA** | 0.01 | 0.92 | 0.96 | 0.03 | 0.43 | 0.72 | 15.97 | 0.22 | 0.67 | 4.44 |
| **NMDS** | 0.01 | 0.92 | 0.95 | 0.01 | 0.43 | 0.72 | 14.19 | 0.22 | 0.63 | 5.22 |
| **MDSbc** | 0.02 | 0.90 | 0.93 | 0.10 | 0.41 | 0.77 | 10.30 | 0.21 | 0.54 | 5.33 |

*This table presents the best DR validation values of the different indices for the Gastric mucosa microbiome dataset. In the table, the high-dimensional (HD) space was included in order to show the performance of the indices in the original high-dimensional space. Moreover, an average of rank performances (denoted as AVG Rank) of the indices values was calculated, where the lowest AVG Rank value represents the algorithm evaluated as the best DR method (as the one that offers the best group separability of the samples) by a higher number of indices.*

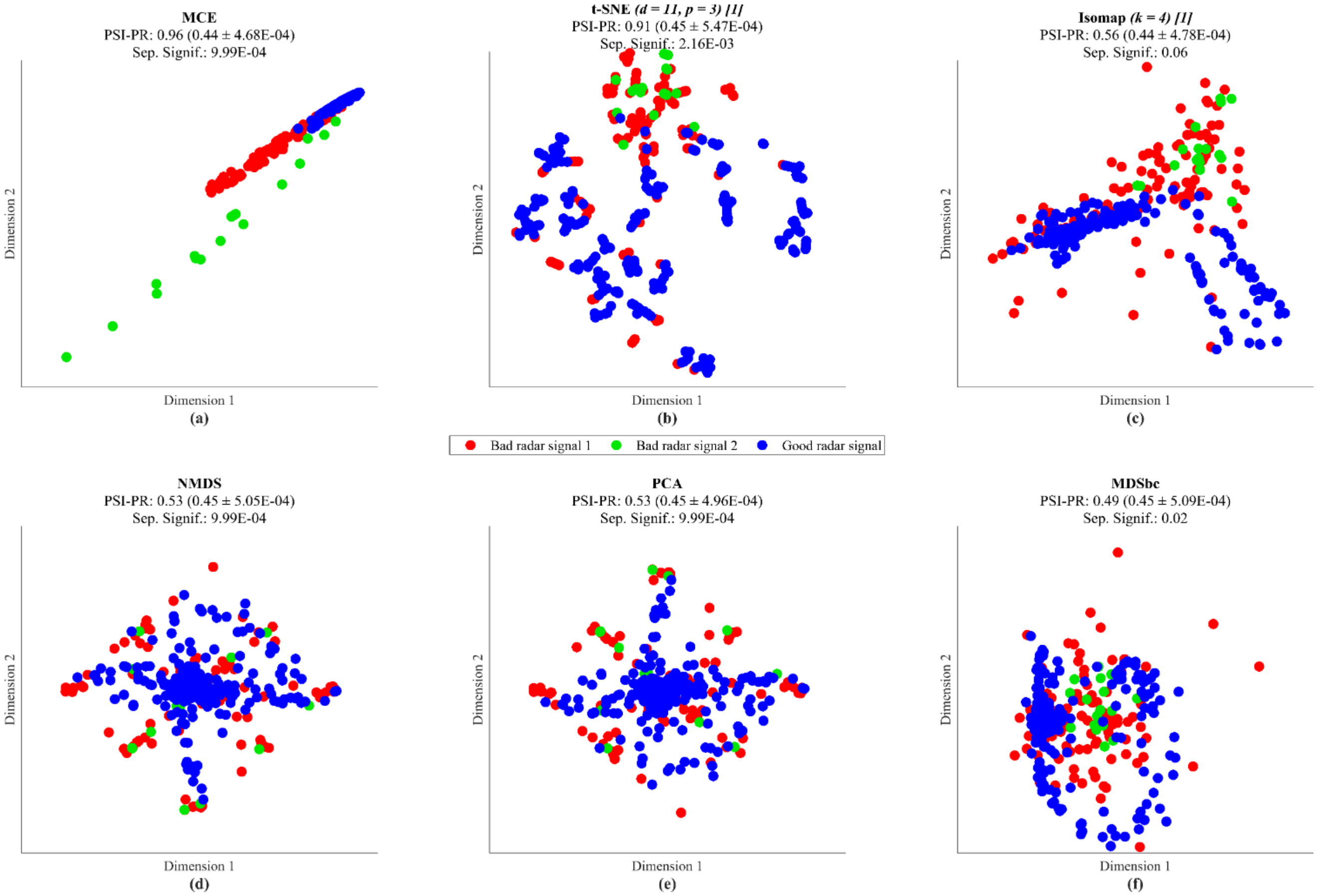

**Fig. 4. PSI-PR of the DR methods applied to Radar signal dataset.**
The DR methods were sorted from the best (top left) to the worst (bottom right) in (a)-(f) according to the PSI-PR validation value. In (b) and (c), the optimal tuning parameter is specified in parentheses next to the title of each DR method, moreover, the digit in brackets represents the number of possible solutions with the same validation value. Furthermore, next to the value returned by this index, it is reported in parentheses the random baseline, which is the mean and the standard error of the PSI-PRs computed by randomly re-shuffling the class labels 1000 times. Finally, we added a p-value which indicates the separability significance of each index in comparison to a respective null model computed by random re-shuffling.

TABLE 4

VALUES OF THE DIFFERENT INDICES OBTAINED FOR RADAR SIGNAL DATASET

| Method | PSIP | PSIROC | PSIPR | DN | DB* | BZ | CH | SH | TH | AVG Rank |
|---|---|---|---|---|---|---|---|---|---|---|
| **MCE** | 1.51E-11 | 0.97 | 0.96 | 1.11E-03 | 0.54 | 0.67 | 317.11 | 0.39 | 0.92 | 1.78 |
| **t-SNE** | 4.16E-09 | 0.91 | 0.91 | 0.04 | 0.42 | 0.69 | 67.50 | 0.17 | 0.92 | 1.89 |
| **Isomap** | 3.35E-04 | 0.75 | 0.56 | 4.79E-03 | 0.29 | 0.58 | 21.40 | 0.09 | 0.83 | 3.89 |
| **HD** | 7.27E-04 | 0.77 | 0.65 | 0.05 | 0.21 | 0.60 | 10.26 | 0.05 | 0.82 | 3.89 |
| **PCA** | 0.01 | 0.69 | 0.53 | 3.07E-03 | 0.30 | 0.60 | 21.52 | 0.03 | 0.78 | 4.44 |
| **NMDS** | 0.01 | 0.68 | 0.53 | 3.07E-03 | 0.22 | 0.60 | 8.67 | 0.02 | 0.66 | 5.78 |
| **MDSbc** | 0.01 | 0.68 | 0.49 | 1.15E-03 | 0.19 | 0.61 | 6.13 | 0.02 | 0.70 | 5.89 |

*This table presents the best DR validation values of the different indices for the Radar signal dataset. In the table, the high-dimensional (HD) space was included in order to show the performance of the indices in the original high-dimensional space. Moreover, an average of rank performances (denoted as AVG Rank) of the indices values was calculated, where the lowest AVG Rank value represents the algorithm evaluated as the best DR method (as the one that offers the best group separability of the samples) by a higher number of indices.*

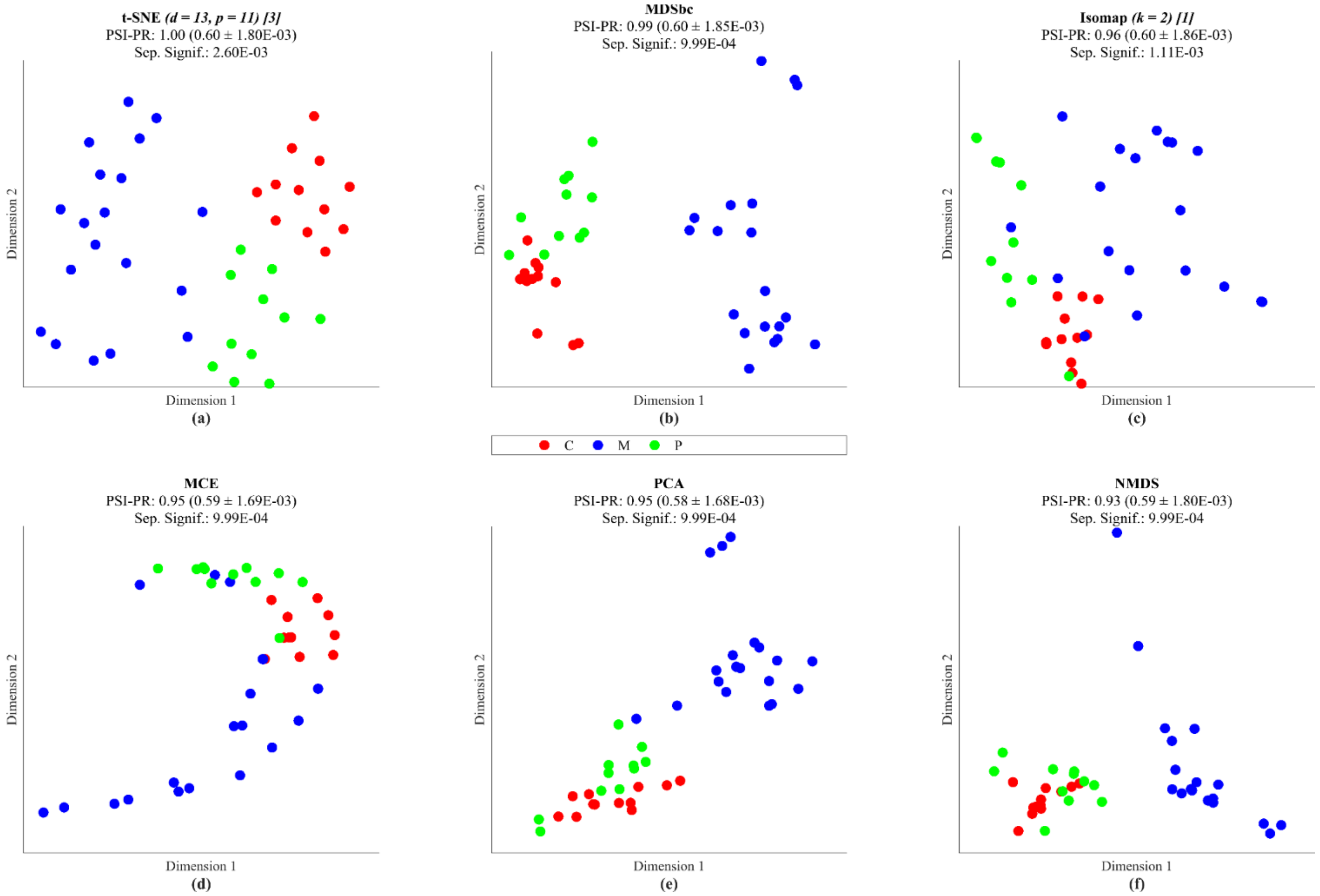

**Fig. 5. PSI-PR of the DR methods applied to Image proteomics dataset.**
The DR methods were sorted from the best (top left) to the worst (bottom right) in (a)-(f) according to the PSI-PR validation value. In (a) and (c), the optimal tuning parameter is specified in parentheses next to the title of each DR method, moreover, the digit in brackets represents the number of possible solutions with the same validation value. Furthermore, next to the value returned by this index, it is reported in parentheses the random baseline, which is the mean and the standard error of the PSI-PRs computed by randomly re-shuffling the class labels 1000 times. Finally, we added a p-value which indicates the separability significance of each index in comparison to a respective null model computed by random re-shuffling.

## TABLE 5

### VALUES OF THE DIFFERENT INDICES OBTAINED FOR IMAGE PROTEOMICS DATASET

| Method | PSIP | PSIROC | PSIPR | DN | DB* | BZ | CH | SH | TH | AVG Rank |
|---|---|---|---|---|---|---|---|---|---|---|
| **t-SNE** | 2.24E-05 | 1.00 | 1.00 | 1.26 | 0.60 | 1.57 | 270.36 | 0.72 | 1.00 | 1.00 |
| **MDSbc** | 6.94E-05 | 0.99 | 0.99 | 0.12 | 0.55 | 0.57 | 64.19 | 0.57 | 0.93 | 2.67 |
| **MCE** | 1.24E-04 | 0.96 | 0.95 | 0.19 | 0.52 | 0.61 | 60.45 | 0.42 | 0.95 | 3.22 |
| **Isomap** | 1.17E-04 | 0.95 | 0.96 | 0.12 | 0.52 | 0.70 | 28.37 | 0.48 | 0.90 | 3.67 |
| **HD** | 2.24E-05 | 1.00 | 1.00 | 0.69 | 0.30 | 0.50 | 5.31 | 0.08 | 0.86 | 4.00 |
| **PCA** | 0.01 | 0.92 | 0.95 | 0.10 | 0.39 | 0.50 | 57.51 | 0.34 | 0.90 | 4.89 |
| **NMDS** | 0.02 | 0.92 | 0.93 | 0.03 | 0.36 | 0.50 | 28.37 | 0.28 | 0.83 | 6.00 |

*This table presents the best DR validation values of the different indices for the Image proteomics dataset. In the table, the high-dimensional (HD) space was included in order to show the performance of the indices in the original high-dimensional space. Moreover, an average of rank performances (denoted as AVG Rank) of the indices values was calculated, where the lowest AVG Rank value represents the algorithm evaluated as the best DR method (as the one that offers the best group separability of the samples) by a higher number of indices.*

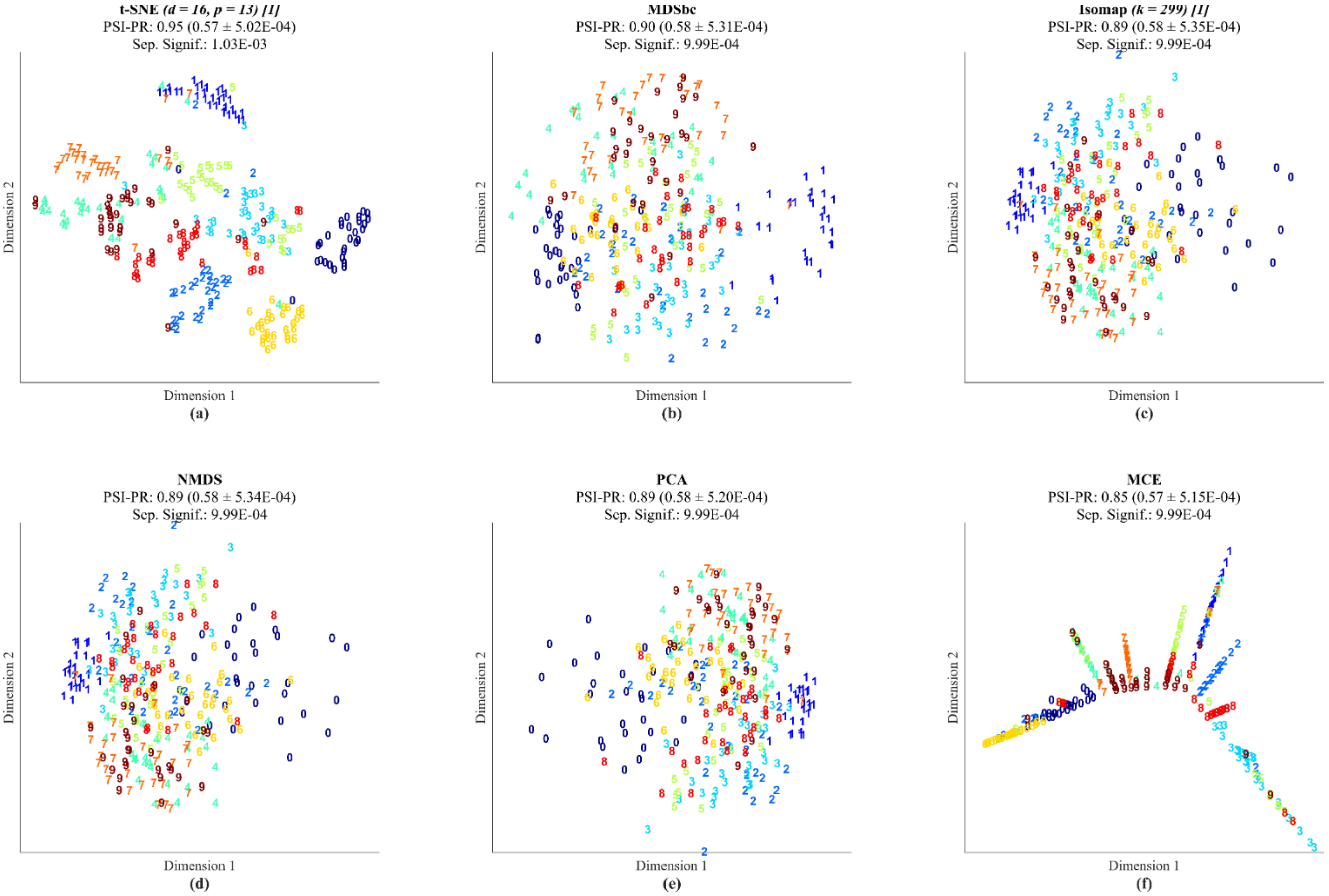

**Fig. 6. PSI-PR of the DR methods applied to MNIST dataset in a two-dimensional (2D) space.**
The DR methods were sorted from the best (top left) to the worst (bottom right) in (a)-(f) according to the PSI-PR validation value. In (a) and (c), the optimal tuning parameter is specified in parentheses next to the title of each DR method, moreover, the digit in brackets represents the number of possible solutions with the same validation value. Furthermore, next to the value returned by this index, it is reported in parentheses the random baseline, which is the mean and the standard error of the PSI-PRs computed by randomly re-shuffling the class labels 1000 times. Finally, we added a p-value which indicates the separability significance of each index in comparison to a respective null model computed by random re-shuffling.

**TABLE 6**

**VALUES OF THE DIFFERENT INDICES OBTAINED FOR MNIST DATASET IN A TWO-DIMENSIONAL (2D) SPACE**

| Method | PSIP | PSIROC | PSIPR | DN | DB* | BZ | CH | SH | TH | AVG Rank |
|---|---|---|---|---|---|---|---|---|---|---|
| **t-SNE** | 1.39E-05 | 0.96 | 0.95 | 0.14 | 0.39 | 0.77 | 130.43 | 0.33 | 0.88 | 1.44 |
| **HD** | 1.08E-09 | 0.99 | 0.99 | 0.50 | 0.24 | 0.66 | 10.22 | 0.07 | 0.82 | 2.22 |
| **Isomap** | 0.01 | 0.90 | 0.89 | 0.04 | 0.23 | 0.50 | 56.57 | -0.06 | 0.46 | 3.89 |
| **MCE** | 0.01 | 0.89 | 0.85 | 0.01 | 0.19 | 0.56 | 85.41 | -0.06 | 0.67 | 4.11 |
| **MDSbc** | 0.02 | 0.90 | 0.90 | 0.01 | 0.18 | 0.66 | 53.16 | -0.04 | 0.40 | 4.33 |
| **PCA** | 0.05 | 0.89 | 0.89 | 0.02 | 0.15 | 0.66 | 44.34 | -0.07 | 0.33 | 5.11 |
| **NMDS** | 0.05 | 0.89 | 0.89 | 4.52E-03 | 0.15 | 0.66 | 44.34 | -0.07 | 0.33 | 5.44 |

*This table presents the best DR validation values of the different indices for the MNIST dataset (2D space). In the table, the high-dimensional (HD) space was included in order to show the performance of the indices in the original high-dimensional space. Moreover, an average of rank performances (denoted as AVG Rank) of the indices values was calculated, where the lowest AVG Rank value represents the algorithm evaluated as the best DR method (as the one that offers the best group separability of the samples) by a higher number of indices.*

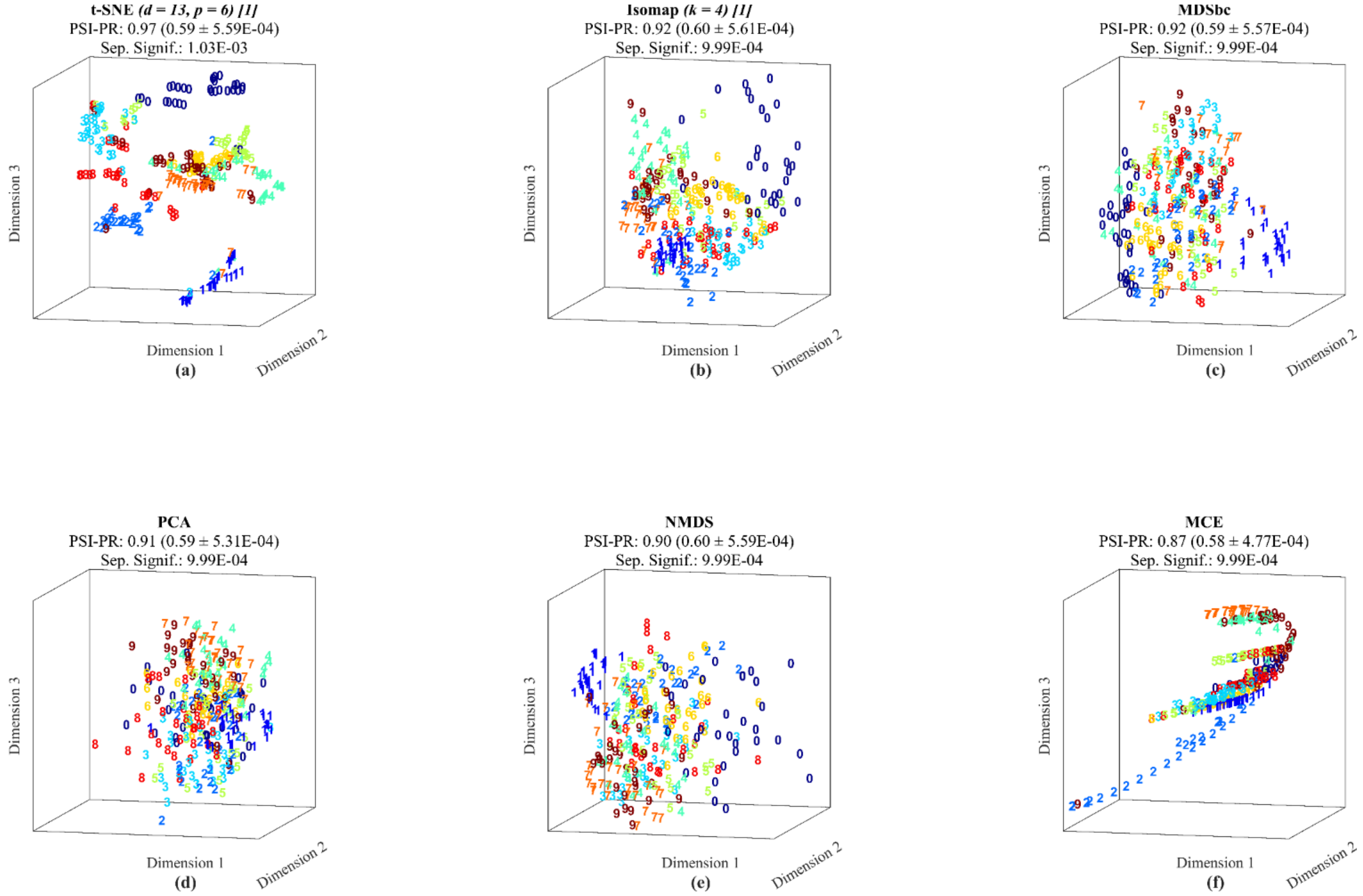

**Fig. 7. PSI-PR of the DR methods applied to MNIST dataset in a three-dimensional (3D) space.**
The DR methods were sorted from the best (top left) to the worst (bottom right) in (a)-(f) according to the PSI-PR validation value. In (a) and (b), the optimal tuning parameter is specified in parentheses next to the title of each DR method, moreover, the digit in brackets represents the number of possible solutions with the same validation value. Furthermore, next to the value returned by this index, it is reported in parentheses the random baseline, which is the mean and the standard error of the PSI-PRs computed by randomly re-shuffling the class labels 1000 times. Finally, we added a p-value which indicates the separability significance of each index in comparison to a respective null model computed by random re-shuffling.

TABLE 7

VALUES OF THE DIFFERENT INDICES OBTAINED FOR MNIST DATASET IN A THREE-DIMENSIONAL (3D) SPACE

| Method | PSIP | PSIROC | PSIPR | DN | DB* | BZ | CH | SH | TH | AVG Rank |
|---|---|---|---|---|---|---|---|---|---|---|
| **t-SNE** | 4.76E-07 | 0.97 | 0.97 | 0.14 | 0.41 | 0.78 | 107.74 | 0.37 | 0.88 | 1.44 |
| **HD** | 1.08E-09 | 0.99 | 0.99 | 0.50 | 0.24 | 0.66 | 10.22 | 0.07 | 0.82 | 2.44 |
| **Isomap** | 3.83E-04 | 0.94 | 0.92 | 0.07 | 0.28 | 0.57 | 55.19 | 0.04 | 0.60 | 3.22 |
| **MDSbc** | 0.01 | 0.92 | 0.92 | 0.05 | 0.20 | 0.66 | 38.53 | -0.01 | 0.49 | 4.56 |
| **PCA** | 3.44E-03 | 0.92 | 0.91 | 0.05 | 0.24 | 0.66 | 35.11 | -0.01 | 0.42 | 4.67 |
| **MCE** | 1.77E-03 | 0.91 | 0.87 | 0.01 | 0.20 | 0.56 | 80.78 | -0.06 | 0.70 | 5.44 |
| **NMDS** | 0.02 | 0.91 | 0.90 | 0.04 | 0.18 | 0.66 | 35.11 | -0.01 | 0.42 | 5.67 |

*This table presents the best DR validation values of the different indices for the MNIST dataset (3D space). In the table, the high-dimensional (HD) space was included in order to show the performance of the indices in the original high-dimensional space. Moreover, an average of rank performances (denoted as AVG Rank) of the indices values was calculated, where the lowest AVG Rank value represents the algorithm evaluated as the best DR method (as the one that offers the best group separability of the samples) by a higher number of indices.*

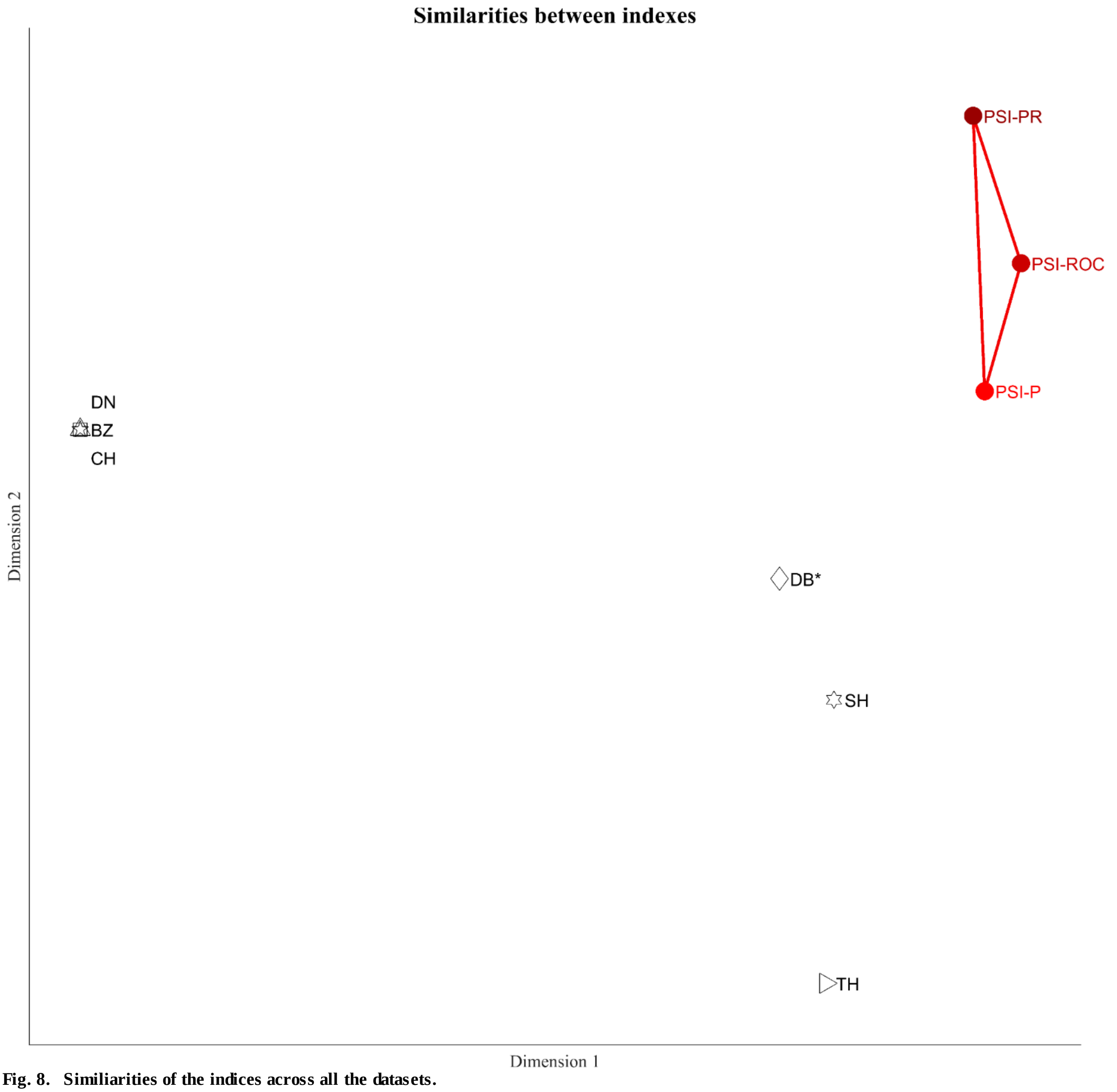

**Fig. 8. Similiarities of the indices across all the datasets.**
We create a matrix by merging row vectors, each of which reports the result of a different separability index in all the datasets for all the DR methods. Then, centered PCA is applied to this matrix (after z-scoring the indices' values for each row, because each index has a different scale) and the first two principal componentsare are plotted in order to visualize the similiarities between the indices. PSIs indices are red marks, CVIs indices are black marks. The PSI triangle (triangle whose vertices are the PSI indices; PSI-P, PSI-ROC, and PSI-PR) is drawn with the aim to show which indices gave comparable results to PSI, because they are projected inside the trinagle. It is evident that all indices are far from the triangle area, therefore the proposed PSI indices represent a novel separated class of indices for evaluation of group separability in a geometrical space.